\documentclass[10pt,twocolumn,letterpaper]{article}

\newcommand{\arxivversion}{1} 

\ifdefined\arxivversion
    \usepackage[pagenumbers]{cvpr} 
\else
    \usepackage[review]{cvpr}      
\fi

\usepackage[dvipsnames]{xcolor}
\usepackage{placeins}

\definecolor{cvprblue}{rgb}{0.21,0.49,0.74}
\usepackage[pagebackref,breaklinks,colorlinks,citecolor=cvprblue]{hyperref}

\def\paperID{12716} 
\def\confName{CVPR}
\def\confYear{2024}

\title{UnifiedVisionGPT: Streamlining Vision-Oriented AI through Generalized Multimodal Framework}

\author{Chris Kelly\thanks{* Authors contributed equally.}\\
Stanford University\\
{\tt\small ckelly24@stanford.edu}
\and
Luhui Hu\footnotemark[1]\\
Seeking AI\\
\and
Cindy Yang\\
University of Washington, Seattle\\
{\tt\small c1ndyy@uw.edu}
\and
Yu Tian\\
Harvard University\\
{\tt\small ytian11@meei.harvard.edu}
\and
Deshun Yang\\
Seeking AI\\
\and
Bang Yang\\
Peking University\\
{\tt\small yangbang@pku.edu.cn}
\and
Zaoshan Huang\\
Seeking AI\\
\and
Zihao Li\\
Seeking AI\\
{\tt\small li981354@seeking.ai}
\and
Yuexian Zou\\
Peking University\\
{\tt\small zouyx@pku.edu.cn}
}

\begin{document}
\maketitle
\begin{abstract}
In the current landscape of artificial intelligence, foundation models serve as the bedrock for advancements in both language and vision domains. OpenAI GPT-4 \cite{gpt4} has emerged as the pinnacle in large language models (LLMs), while the computer vision (CV) domain boasts a plethora of state-of-the-art (SOTA) models such as Meta's SAM \cite{sam} and DINO \cite{dinov2,darcet}, and YOLOS \cite{yoloreview,yolo,yolonas}. However, the financial and computational burdens of training new models from scratch remain a significant barrier to progress. In response to this challenge, we introduce UnifiedVisionGPT, a novel framework designed to consolidate and automate the integration of SOTA vision models, thereby facilitating the development of vision-oriented AI. UnifiedVisionGPT distinguishes itself through four key features: (1) provides a versatile multimodal framework adaptable to a wide range of applications, building upon the strengths of multimodal foundation models; (2) seamlessly integrates various SOTA vision models to create a comprehensive multimodal platform, capitalizing on the best components of each model; (3) prioritizes vision-oriented AI, ensuring a more rapid progression in the CV domain compared to the current trajectory of LLMs; and (4) introduces automation in the selection of SOTA vision models, generating optimal results based on diverse multimodal inputs such as text prompts and images. This paper outlines the architecture and capabilities of UnifiedVisionGPT, demonstrating its potential to revolutionize the field of computer vision through enhanced efficiency, versatility, generalization, and performance. Our implementation, along with the unified multimodal framework and comprehensive dataset, is made publicly available at https://github.com/LHBuilder/SA-Segment-Anything.
\end{abstract}    
\section{Introduction}
\label{sec:intro}

In the rapidly evolving era defined by generative AI (GAI), two trends have seemed to rise above the rest: Large-Language Models and Computer Vision large-scale models. Although models like GPT-4 have established a remarkable benchmark for LLMs, the field of multimodal CV models is an ever-changing frontier, full of potential. \\
\\
We look to tap into this potential through its vision-oriented multimodal framework built in UnifiedVisionGPT. Unlike a traditional LLM or vision foundation model, UnifiedVisionGPT integrates multiple large-scale models, some built atop the most advanced foundation models available. One of these models is Meta’s Segment Anything Model (SAM) which has the ability to segment or “cut out” objects within an image. The other main foundation models are the latest YOLO (You Only Look Once) models (e.g., YOLO-NAS and YOLOv8) that can rapidly detect objects within an image. There are more SOTA vision foundation models, such as Meta's DINO \cite{dinov2,darcet} and Detectron2 \cite{detectron2}, SAM's variants (e.g., FastSAM and MobileSAM) \cite{faster,fast}, and OpenAI's DALL-E \cite{zeroshot,dalle} and CLIP \cite{clip}. UnifiedVisionGPT leverages these models and the state-of-the-art features that set them apart from each other to accelerate CV development. \\
\\
UnifiedVisionGPT serves multiple purposes that align with the current and future needs of the AI community. By providing a unified framework for multimodal applications, this project will help accelerate the development of vision-oriented AI and bridge the gap between the status quo of LLMs and the emergent CV multimodal paradigm. This paper will explore the capabilities, methodologies, and potential future applications of this technology.\\
\\
Through an in-depth examination of the UnifiedVisionGPT architecture, this paper aims to elucidate how the project provides a glimpse into the future where AI can see, interpret, and engage with the world in a manner reminiscent of human intelligence.\\
\\
UnifiedVisionGPT leverages many SOTA CV models, for instance, YOLOv8 model and Meta SAM model. Both of these are highly effective in their own right. YOLO model excels in object detection which means that it can rapidly identify objects in an image and classify them with a label. SAM model can segment any object which makes it useful for many different images. SAM will segment an object by creating a mask that highlights the entirety of the object. Although both of these models accomplish similar tasks on their own, a more powerful model emerges when they are put together with intelligence. \\
\\
For example, the SAM model can identify and segment an object on its own, but the task can be achieved even faster when the two models work together. The YOLO model can be used for the detection of the object and then once the object is found, SAM can be called upon to create a mask for the object.
Furthermore, this framework is especially useful for images that call for instance segmentation, where distinct objects of the same class need to be differentiated from each other through different colored masks. \\
\\
To help understand each of these foundational models and how they work together in our unified framework, consider these three images:\\
\begin{figure}[htbp]
    \centering
    \includegraphics[width=0.5\linewidth]{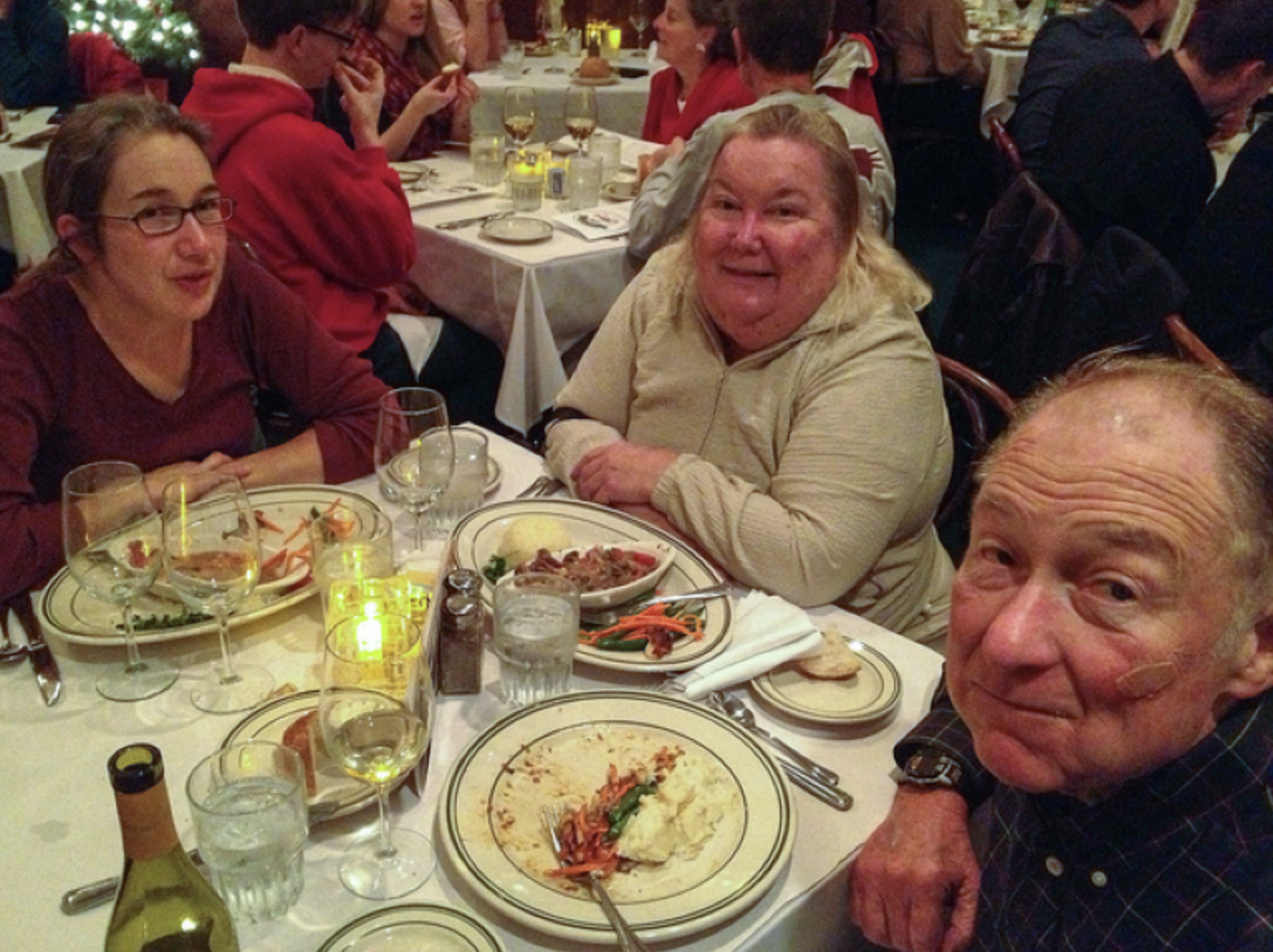}
    \caption{Image before processing}
    \label{fig:dinner_no_edit}
\end{figure}
\begin{figure}[htbp]
    \begin{minipage}{0.45\linewidth}
        \includegraphics[width=\linewidth]{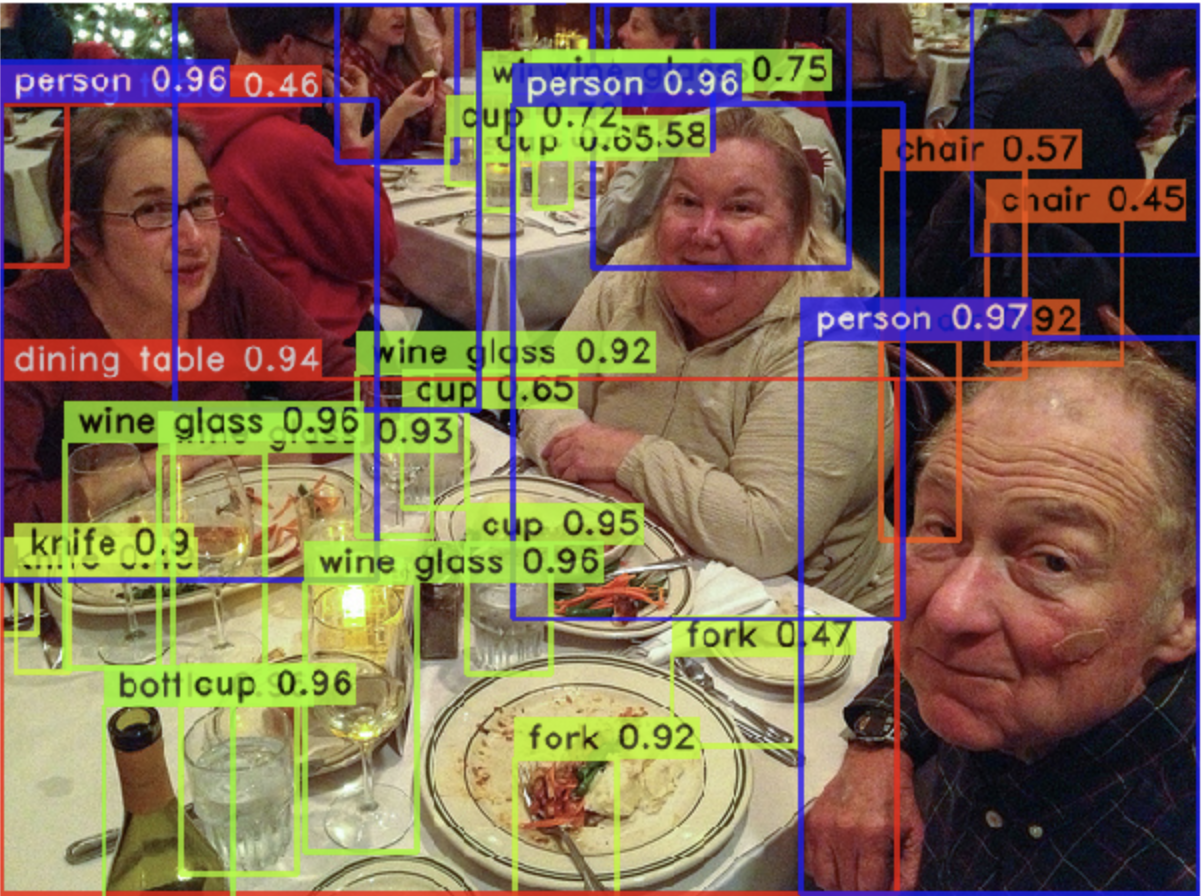}
        \caption{Image with YOLO detection}
        \label{fig:dinner_segmented}
    \end{minipage}
    \hfill
    \begin{minipage}{0.45\linewidth}
        \includegraphics[width=\linewidth]{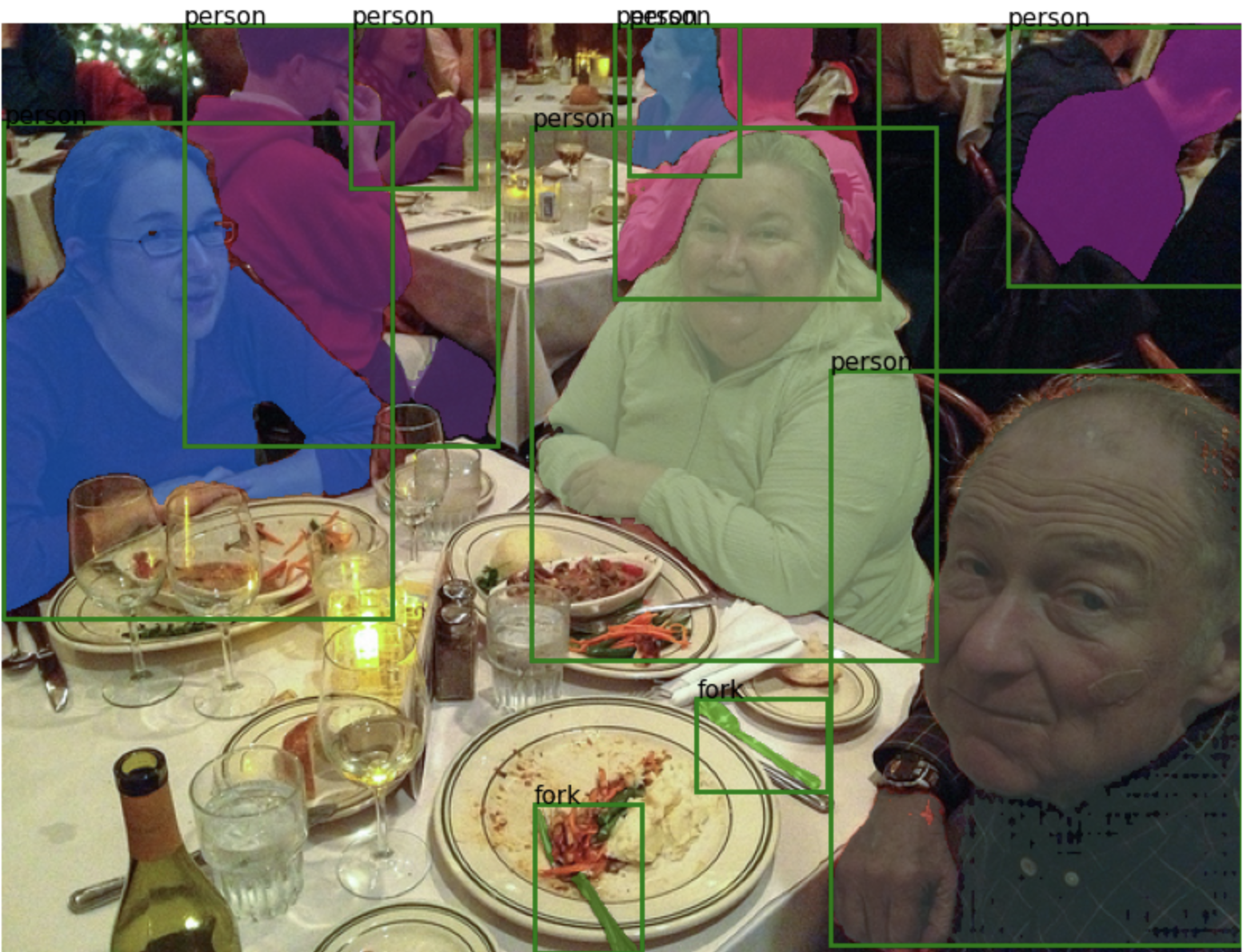}
        \caption{Image with YOLO detection and SAM masking}
        \label{fig:dinner_mask_and_seg}
    \end{minipage}   
\end{figure}
\\
Figure 1 shows the image before any sort of computer vision framework is applied to it. Figure 2 depicts the image after the YOLO module has been applied. In this example, every individual object is detected and given a label. Because there are many different, unique objects in this image, there are many boxes and labels that overlap, which can make it difficult to differentiate between one object from another. That is where the power of the SAM module comes into play. In Figure 3, a specific input has been given that asks to find instances of "fork" and "person". These objects retain their detection from the previous image, but now are also given unique masks with the help of the SAM module. The importance of instance segmentation is apparent as the individuals sitting at the close table are all given different colored masks. \\
\begin{figure}[htbp]
    \centering
    \includegraphics[width=1\linewidth]{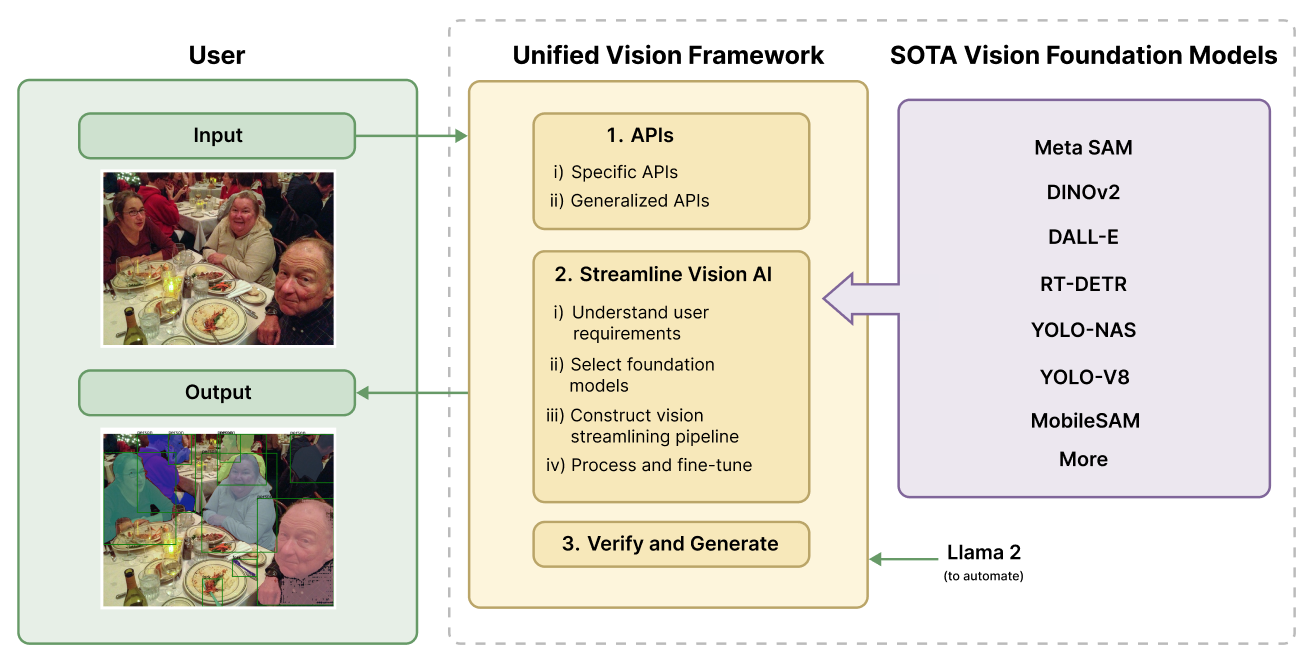}
    \caption{UnifiedVisionGPT Generalized Vision Framework}
    \label{fig:unified_vision_generalized}
\end{figure}
\\
This unified framework has incredibly high upside and numerous potential applications through integrating with an open-source LLM (Meta's Llama 2) \cite{llama2}. UnifiedVisionGPT employs this LLM as a sort of director that can interpret the user's requests and act accordingly. Depending on what the user requests, a certain CV model might be called over another or both of them together. The important factor lies in the customization of the user’s request that will be met through UnifiedVisionGPT’s use of an LLM and its unified framework. This will allow a user to make custom requests that can be interpreted by the LLM and turned into action items that UnifiedVisionGPT can manage.\\
\\
The controlflow of our unified framework and its connection with a LLM can be broken down into a few different tasks: \\
\begin{enumerate}
    \item \textbf{Vision Pre-processing:} In this step, the LLM interprets the user's request and breaks down the instructions into smaller action items. The original image is also uploaded to the unified framework.
    \item \textbf{Foundation Model Selection:} The appropriate foundation model(s) is selected depending on the individual action items.
    \item \textbf{Execution:} Foundation model executed on appropriate objects within the image.
    \item \textbf{Post Processing and Integration:} Edited images containing segmentation and/or masks returned to user through the framework
\end{enumerate}
For instance, consider two distinct images as an input unlike simply only one image. A user might make this request: \textit{“Find dogs and lemons in the images and then highlight them only”}. UnifiedVisionGPT will have a LLM interpret this request as part of Task 1: The Vision Preprocessing Stage. Following this stage, the actionable items should be broken down into requests such as this:\\
\begin{enumerate}[label=\Alph*.]
    \item Locate dogs
    \item Highlight dogs by segment "mask"
    \item Locate lemons
    \item Highlight lemons by segment "mask"
    \item Integrate all images together (optional)
\end{enumerate}
\smallskip
After the instructions have been interpreted, the best-suited foundation model will be selected. In this case, request 1 calls for a simple object detection of all dogs, so the YOLO-NAS model (or YOLOv8) will be called upon. Request 2 asks for the dogs to be highlighted, which means that the masking abilities of SAM will be needed. However, SAM can leverage the use of YOLO-NAS to save itself some time; YOLO-NAS has already identified the location of the dog and now SAM can use that information to quickly create a mask for the animal. Request 3 and 4 will repeat as the above. \\

\begin{figure}[htbp]
    \centering
    \includegraphics[width=1\linewidth]{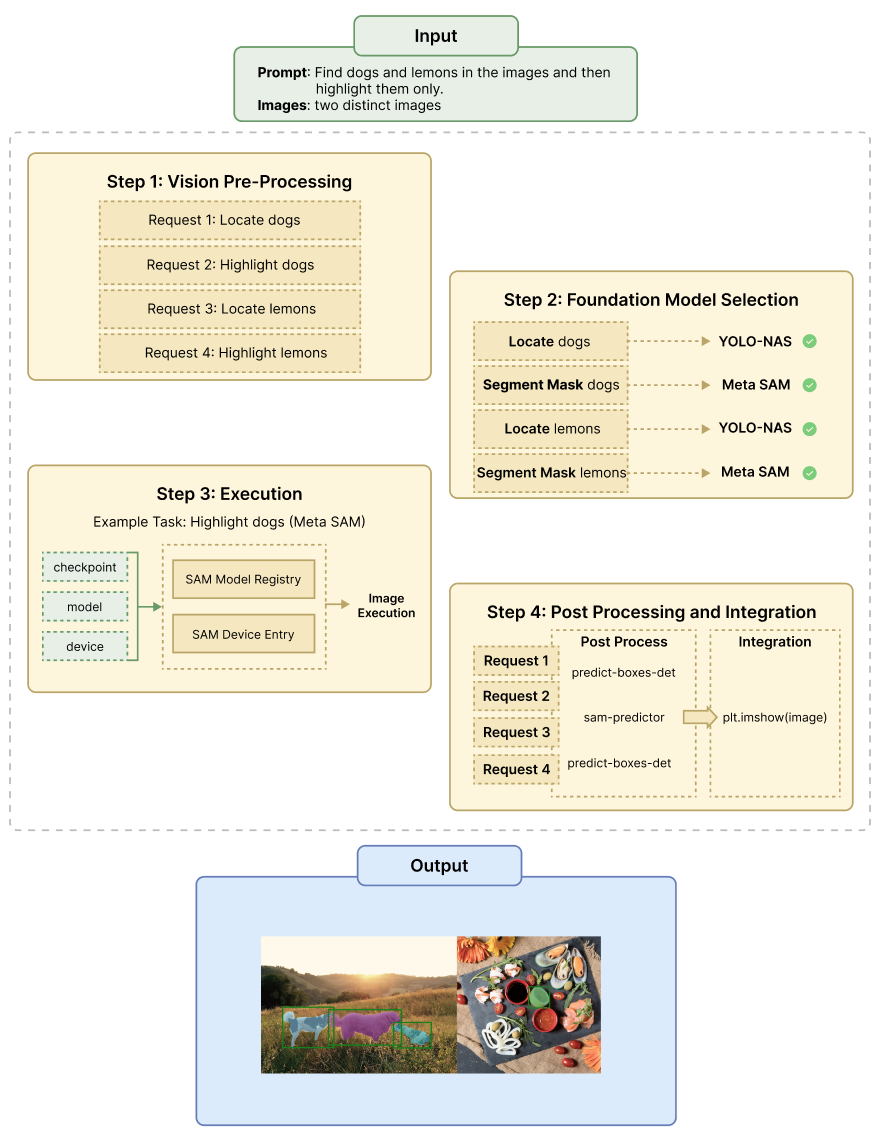}
    \caption{Overview of UnifiedVisionGPT Generalized Framework}
    \label{fig:unified_vision_overview}
\end{figure}

\noindent This is just an example. UnifiedVisionGPT is a multimodal framework to take text prompts and images and other vision files as inputs and then streamlines the vision process through integrating the SOTA vision foundation models.
\section{Related Works}
\label{sec:related_works}

The integration of state-of-the-art (SOTA) computer vision (CV) technologies and large language models (LLMs) has become increasingly prevalent in the field of artificial intelligence. UnifiedVisionGPT uniquely contributes to this domain by synergizing SOTA vision models, such as YOLO \cite{yolo} and Meta SAM \cite{sam}, into a comprehensive multimodal framework. \\
\\
In terms of bridging frameworks with LLMs, HuggingGPT \cite{hugginggpt} emerges as a pertinent reference, connecting the extensive model repository HuggingFace \cite{huggingface} to LLMs like ChatGPT. Although this approach aligns with UnifiedVisionGPT in facilitating interactions with LLMs, UnifiedVisionGPT distinguishes itself by prioritizing a seamless integration and automation of SOTA vision models, thereby fostering a robust and vision-focused AI ecosystem.\\
\\
Grounded SAM \cite{gsam} extends the capabilities of Meta SAM through training with language instructions, enhancing its object segmentation proficiency. This innovation resonates with UnifiedVisionGPT’s objective of harnessing and amplifying the unique capabilities of existing models to advance vision-oriented AI.\\
\\
MiniGPT-4 \cite{minigpt} represents another stride towards integrating visual and linguistic modalities, aligning a visual encoder with the advanced LLM, Vicuna. Although there are parallels in modality integration between MiniGPT-4 and UnifiedVisionGPT, UnifiedVisionGPT stands out by offering a holistic integration and automation of various SOTA vision models, ensuring versatility and peak performance across diverse applications.\\
\\
The work on VoxPoser \cite{voxposer} serves as a related endeavor to UnifiedVisionGPT in integrating Large Language Models (LLMs) with robotic manipulation tasks. VoxPoser uniquely synthesizes robot trajectories and constructs 3D value maps based on natural language instructions, highlighting its strength in dealing with a wide array of manipulation tasks. UnifiedVisionGPT, on the other hand, extends the application of LLMs beyond manipulation, aiming to create a generalized and automated framework for various vision-oriented tasks. Although they share common ground in leveraging LLMs for robotic methods, each framework carves out its niche, contributing uniquely to the intersection of language models and vision-based robotic tasks.\\
\\
In "Physically Grounded Vision-Language Models for Robotic Manipulation" \cite{pgvlm}, the authors address the limitations of current Vision-Language Models (VLMs) in understanding physical concepts crucial for robotic manipulation. They introduce a new dataset, PhysObjects, to enhance the VLM’s comprehension of these concepts, resulting in improved robotic planning performance. While UnifiedVisionGPT focuses on creating a unified and automated framework for a variety of vision-based tasks using Large Language Models, \cite{pgvlm} emphasizes physically-grounding VLMs to augment their utility in robotic manipulation tasks. Both works underscore the significance of integrating language models with visual perception for robotic applications, albeit with different focuses and methodologies.\\
\\
Foundation models and GPTs grow so fast. There are other related works, including Visual ChatGPT, Flamingo, VLMo, VIOLET, and more \cite{visual,vlmo,violet,flamingo}. They mainly focused on ChatGPT integration or transformer-based solutions. UnifiedVisionGPT differentiates them with its generalized multimodal framework. Its own framework can fine-tune the LLM \cite{llama} for its specific streamlining purpose. \cite{llmzs}\\
\\
In conclusion, despite the presence of related works in CV and LLM integration, UnifiedVisionGPT still offers unique capabilities. It does so by unifying and automating the capabilities of SOTA vision models, optimizing multimodal interactions, and delivering a streamlined and efficient user experience.
\section{UnifiedVisionGPT Framework}

UnifiedVisionGPT operates as a cooperative platform designed to tackle object detection and image process through AI tasks, harnessing the capabilities of an LLM in tandem with an array of expert models hailing from the machine learning communities. The process unfolds across four key tasks: vision pre-processing, foundation model selection, execution, and integration and post processing. When confronted with a user's request, UnifiedVisionGPT initiates an automated deployment of the complete workflow. This orchestrates the collaboration and utilization of expert models, such as the YOLO model and the SAM model, to successfully accomplish the designated objectives set up by the user. \\
\\
Establishing a connection between UnifiedVisionGPT and a SFT (supervised fine-tuning) LLM opens up a realm of possibilities that could potentially lead to the ultimate customization of user requests. This synergy between UnifiedVisionGPT and an LLM creates a dynamic ecosystem where the linguistic and reasoning capabilities of the LLM can be seamlessly integrated with the image processing and understanding capability of UnifiedVisionGPT. When a user submits a request, the LLM can first parse the natural language input, extracting a number of details, context, and intent. Simultaneously, UnifiedVisionGPT can analyze any accompanying images or visual data linked to the request. The LLM then arranges a collaboration between these two components, effectively translating the user's request into object analysis and recognition within the image. This connection paves the way for a highly adaptive and context-aware system, capable of adapting its responses to the user's preferences, language nuances, and the specific content of the visual data provided. As the connection between UnifiedVisionGPT and future LLMs evolves, the potential for ultimate customization of user requests becomes increasingly tangible, such as performing specific tasks relating to object recognition, scene understanding, or even creative image generation, tailored to the user's unique needs.\\
\\
UnifiedVisionGPT has four main components: 1. APIs, 2. Streamline Vision AI, 3. Verify and Generate, and 4. Fine Tuning. Please refer to Figure 4 for the first 3 components.
\begin{enumerate}
    \item \textbf{APIs:} UnifiedVisionGPT supports two types of APIs: specific APIs and generalized APIs. The specific APIs can be some simple or standard APIs for some common vision AI operations, for example, labelObjects(\textless object name \textgreater, \textless image location \textgreater). The generalized APIs are flexible for inputs: a text prompt for instruction and a list of images or videos. The prompt can control and instruct the operations on images or videos.
    \item \textbf{Streamline Vision AI:} UnifiedVisionGPT has intelligent logic to automate the process of vision AI based on Llama 2 and the integrated SOTA vision foundation models.
    \item \textbf{Verify and Generate:} UnifiedVisionGPT is unique to verify the results against the inputs for the best results. It will retry if it detects something wrong. For example, if it chose a wrong foundation model at the first place, it would retry to correct. The generation step is an additional step to use other vision tools, such as OpenCV and GAI (Generative AI) to generate based on the API requirements or the input prompts. 
    \item \textbf{Fine Tune:} Llama 2 is Meta's open-source LLM, which is one of the best open source LLMs. But it still has limits in the specific domains and streamlining generalization. So UnifiedVisionGPT has its dedicated vector DB and historical vision-related dataset to fine tune (or supervised fine-tuning) the Llama 2 for a better LLM. 
\end{enumerate}

UnifiedVisionGPT is an open framework from the APIs to the internal streamlining logic. The goal of UnifiedVisionGPT aims to provide a generalized multimodal framework for streamlining versatile vision AI.
\section{Method}

The UnifiedVisionGPT method is a novel multimodal framework designed to enhance vision-oriented AI capabilities. UnifiedVisionGPT combines the strengths of Large Language Models (LLMs) and vision processing techniques, adopting a zero-shot learning approach to generalize and automate a variety of vision tasks based on natural language instructions.

\subsection{Problem Formulation}
Consider a scenario where we are given a natural language instruction \( L \) that describes a specific vision task. The goal of UnifiedVisionGPT is to interpret \( L \) and translate it into a set of visual processing tasks \( T = \{t_1, t_2, \ldots, t_n\} \), where each \( t_i \) represents an individual operation such as object recognition, image segmentation, or feature extraction. The core challenge is to ensure that the interpretation of \( L \) accurately reflects the intended visual task, aligning the output with the user's expectations.

The problem can be mathematically formulated as:

\begin{equation}
    \min_{T} \mathcal{L}(f(L), T) + \mathcal{R}(T)
    \label{eq: important}
\end{equation}

where \( f(L) \) denotes the feature representation of the natural language instruction, \( T \) is the set of generated visual tasks, \( \mathcal{L} \) is a loss function measuring the discrepancy between the generated tasks and the intended visual outcomes, and \( \mathcal{R} \) is a regularization term ensuring the tasks are well-defined and executable.

\subsection{Language-Grounded Visual Task Generation}
UnifiedVisionGPT utilizes an advanced LLM to interpret natural language instructions and generate a corresponding set of visual tasks. The LLM is trained to understand and contextualize language, producing a semantic representation that guides the generation of \( T \). This ensures that the visual tasks are firmly grounded in the linguistic context provided by the user, facilitating accurate and relevant task generation.

\subsection{Zero-Shot Generalization for Vision Task Automation}
UnifiedVisionGPT is designed to generalize across a wide array of vision tasks, utilizing a zero-shot learning approach to handle novel scenarios and instructions. The LLM's extensive pre-training allows it to draw on a broad knowledge base, enabling the generation of visual tasks even in the absence of task-specific training data. This capacity for generalization ensures that UnifiedVisionGPT can automate vision task generation across various contexts and applications, showcasing a robotic level of efficiency and adaptability.

\subsection{Joint Optimization for Coherent Task Execution}
To guarantee that the generated visual tasks are not only linguistically aligned but also lead to successful execution, UnifiedVisionGPT employs a joint optimization strategy. This approach considers both the semantic congruence between \( L \) and \( T \) and the practical feasibility of the visual tasks. Through this comprehensive optimization, UnifiedVisionGPT ensures coherent and effective task execution, aligning the visual output with the user's intent.

In summary, UnifiedVisionGPT introduces an intelligent and robotic methodology for processing and generating visual tasks, seamlessly integrating natural language understanding, visual task generation, and zero-shot learning. This unified approach facilitates intuitive and efficient interactions between users and vision-oriented AI systems, furthering the field of automated and generalized vision processing.
\section{Experiments}

UnifiedVisionGPT can streamline vision tasks by integrating state-of-the-art (SOTA) vision foundation models. In this section, we demonstrate its capabilities through various experiments, showcasing its multimodal framework that accepts both text prompts and images or videos.\\
\\
Here we will use UnifiedVisionGPT generalized API with different prompts and images to see different results.
\\

\noindent \textbf{Case 1:} Given the prompt "find the guitar and segment it" and an image:
\begin{figure}[htbp]
    \centering
    \begin{minipage}{0.45\linewidth}
        \includegraphics[width=\linewidth]{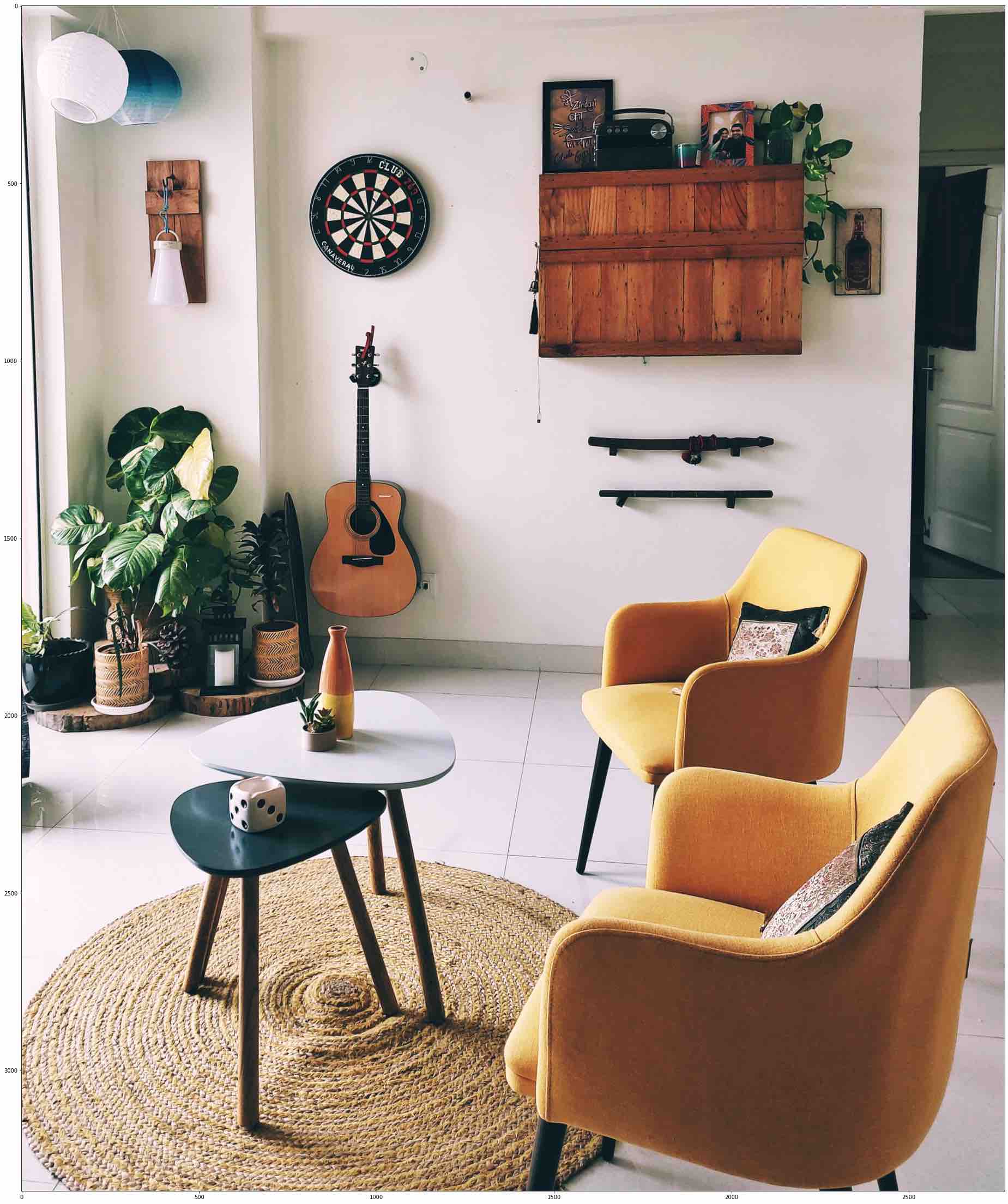}
        \caption{Original image}
        \label{fig:guitar_src}
    \end{minipage}
    \hfill
    \begin{minipage}{0.45\linewidth}
        \includegraphics[width=\linewidth]{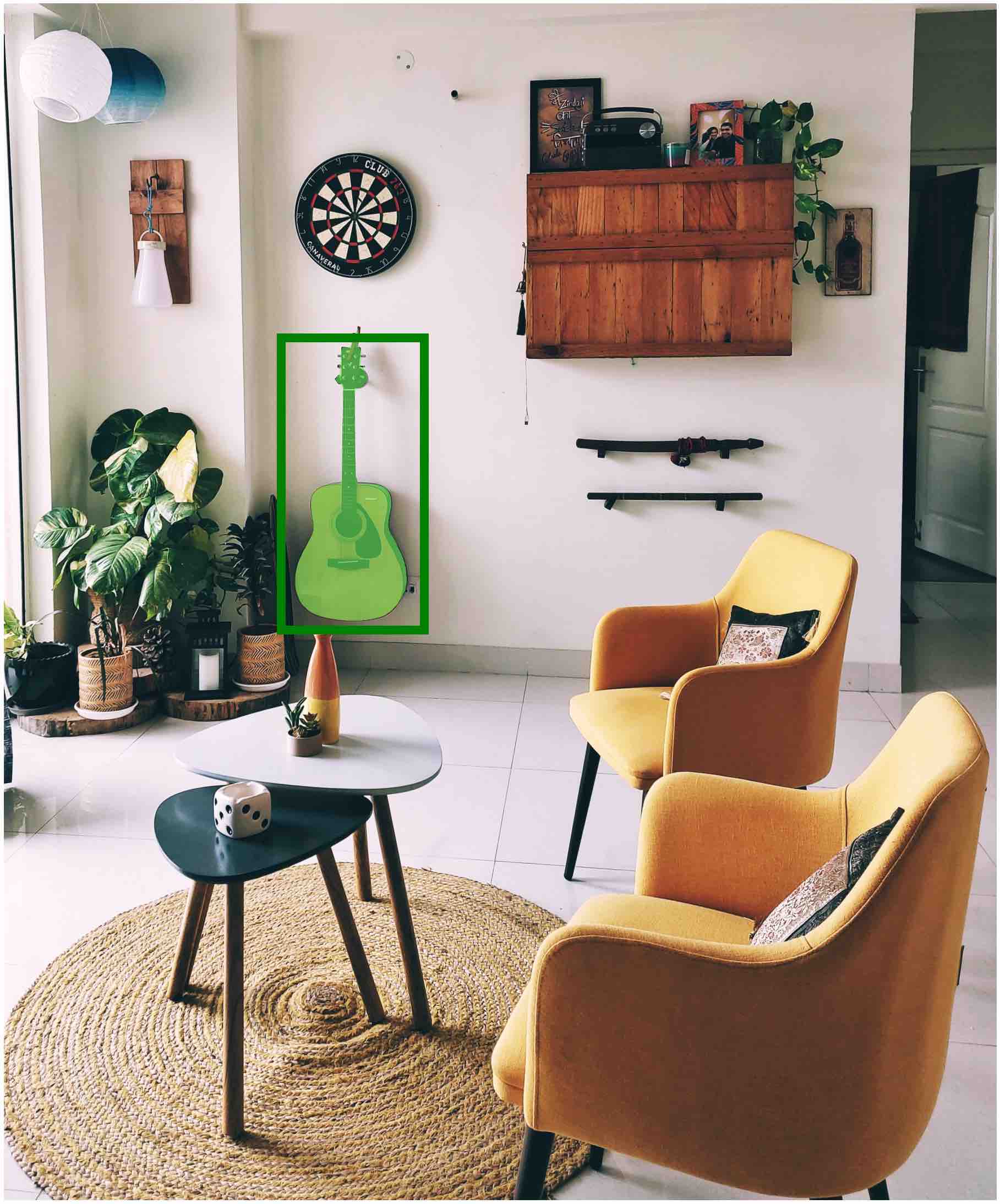}
        \caption{Processed image}
        \label{fig:guitar_res}
    \end{minipage}   
\end{figure}
\\
\newpage 
\FloatBarrier
\noindent \textbf{Case 2:} Given the prompt "find the yellow flower and segment it" and an image:
\begin{figure}[htbp]
    \centering
    \begin{minipage}{0.45\linewidth}
        \includegraphics[width=\linewidth]{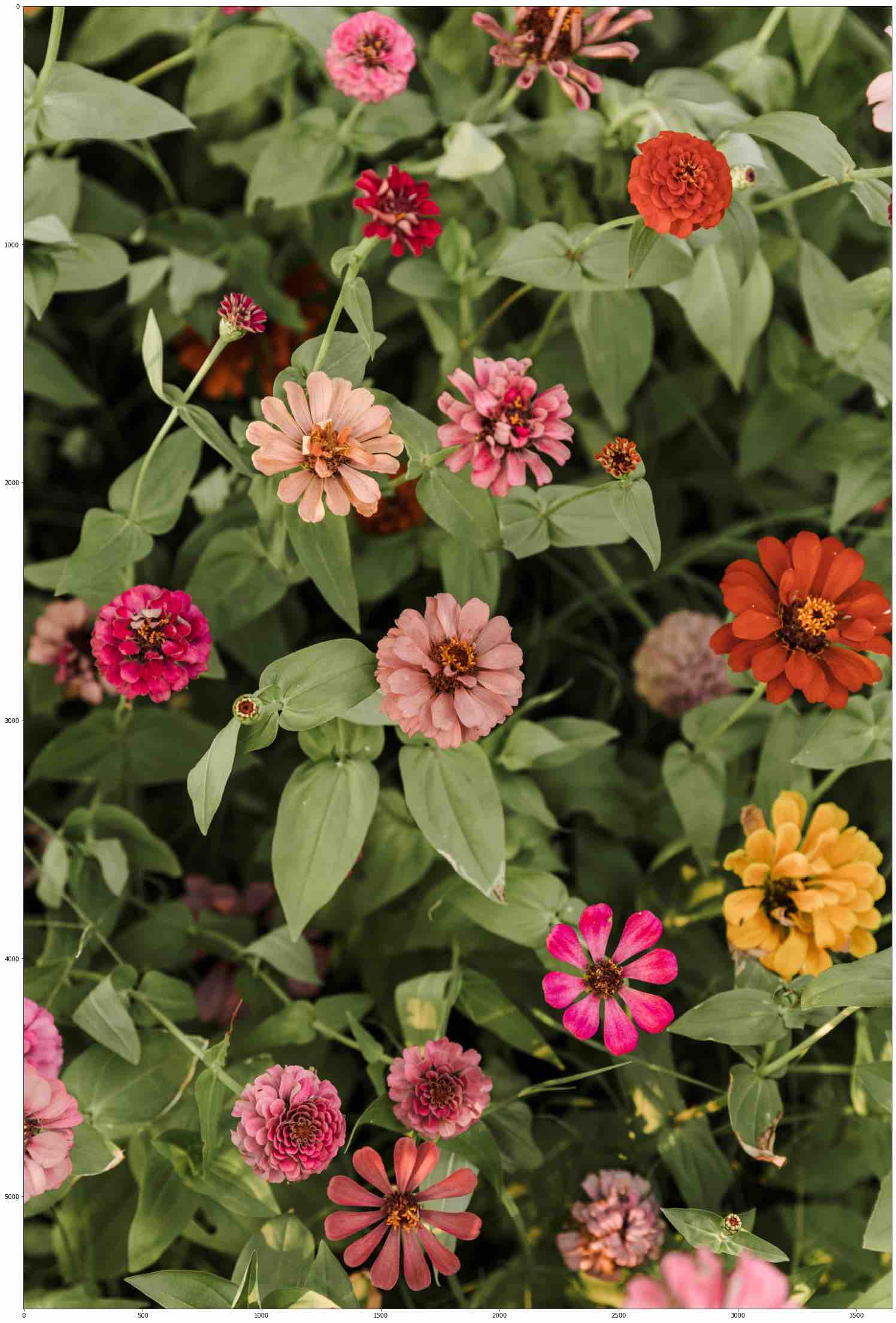}
        \caption{Original image}
        \label{fig:yellow_src}
    \end{minipage}
    \hfill
    \begin{minipage}{0.45\linewidth}
        \includegraphics[width=\linewidth]{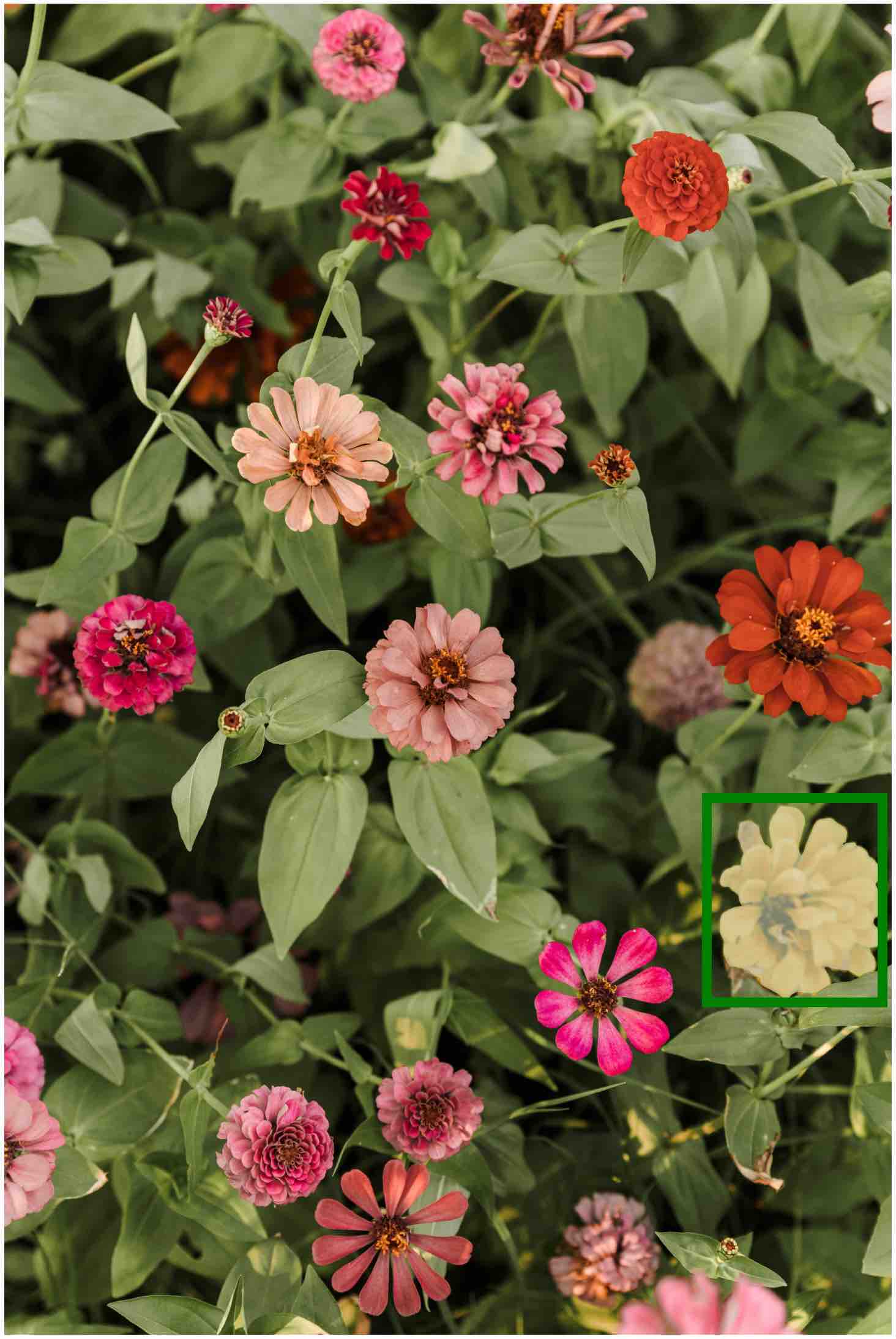}
        \caption{Processed image}
        \label{fig:yellow_res}
    \end{minipage}   
\end{figure}
\\
\FloatBarrier
\noindent \textbf{Case 3:} Given the prompt "find an animal and mask it" and an image:\\
\begin{figure}[htbp]
    \centering
    \begin{minipage}{0.45\linewidth}
        \includegraphics[width=\linewidth]{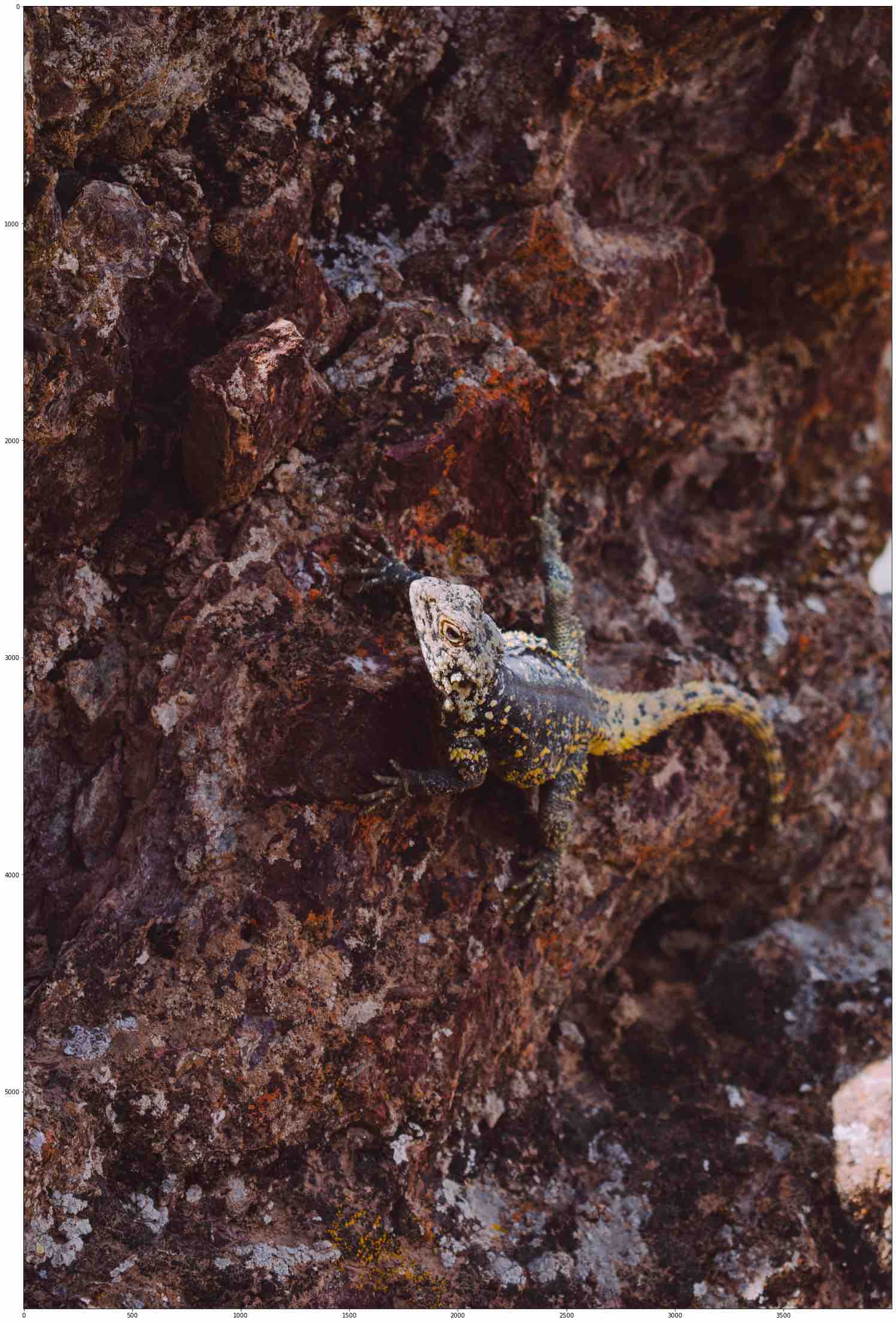}
        \caption{Original image}
        \label{fig:animal_src}
    \end{minipage}
    \hfill
    \begin{minipage}{0.45\linewidth}
        \includegraphics[width=\linewidth]{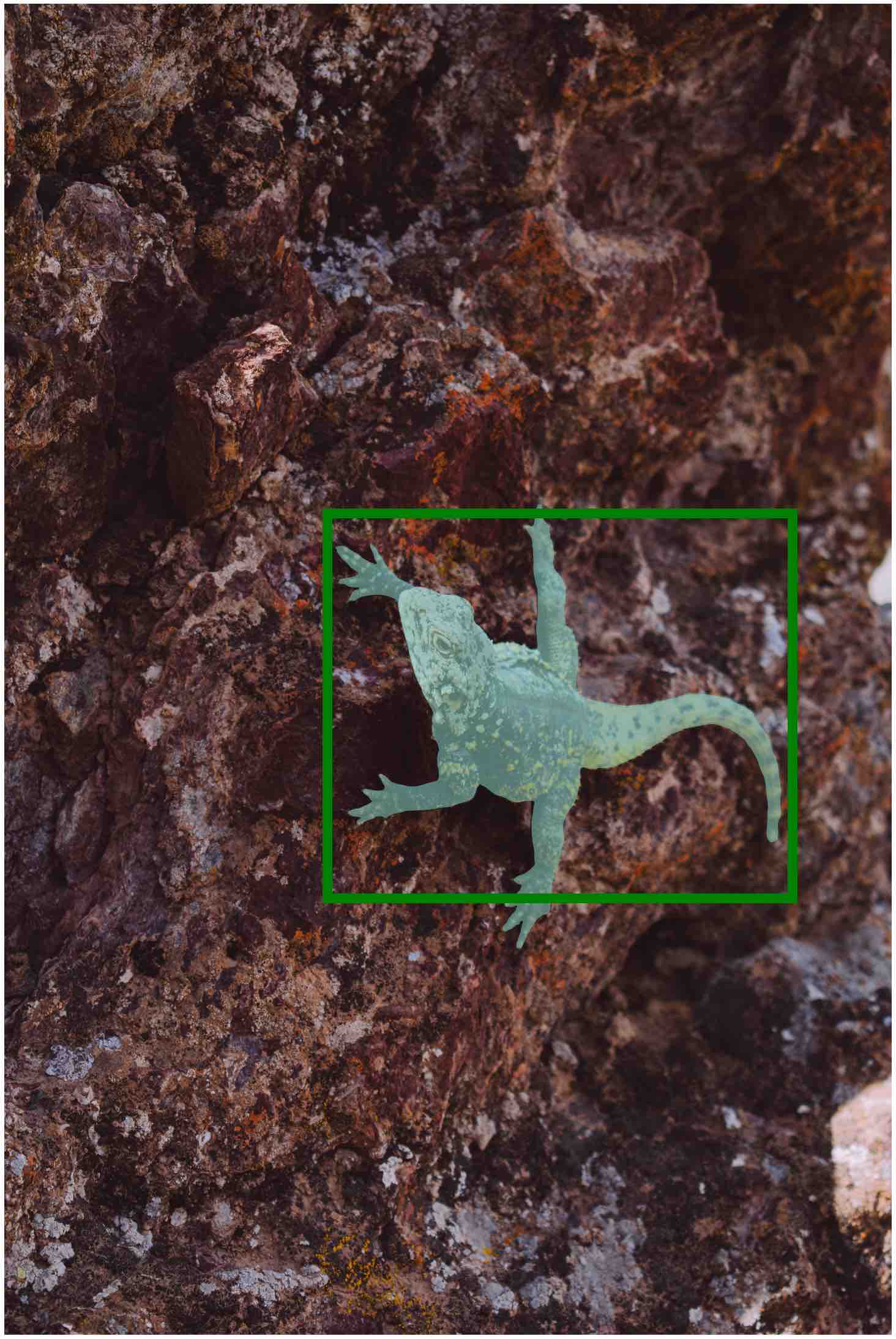}
        \caption{Processed image}
        \label{fig:animal_res}
    \end{minipage}   
\end{figure}
\\
\newpage
\FloatBarrier
\noindent \textbf{Case 4:} Given the prompt "detect frog and then highlight it with masking" and an image:\\
\begin{figure}[htbp]
    \centering
    \begin{minipage}{0.45\linewidth}
        \includegraphics[width=\linewidth]{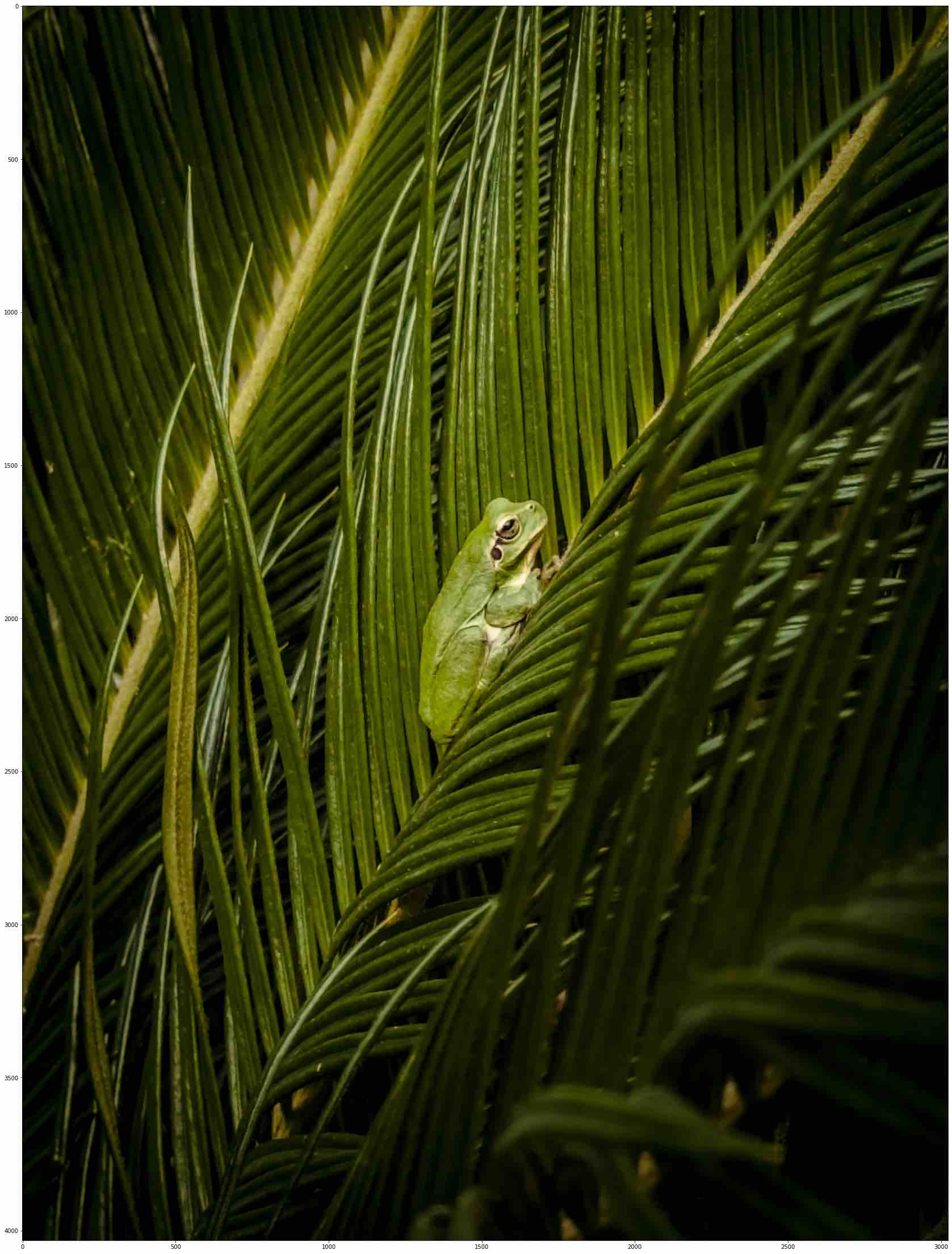}
        \caption{Original image}
        \label{fig:frog_src}
    \end{minipage}
    \hfill
    \begin{minipage}{0.45\linewidth}
        \includegraphics[width=\linewidth]{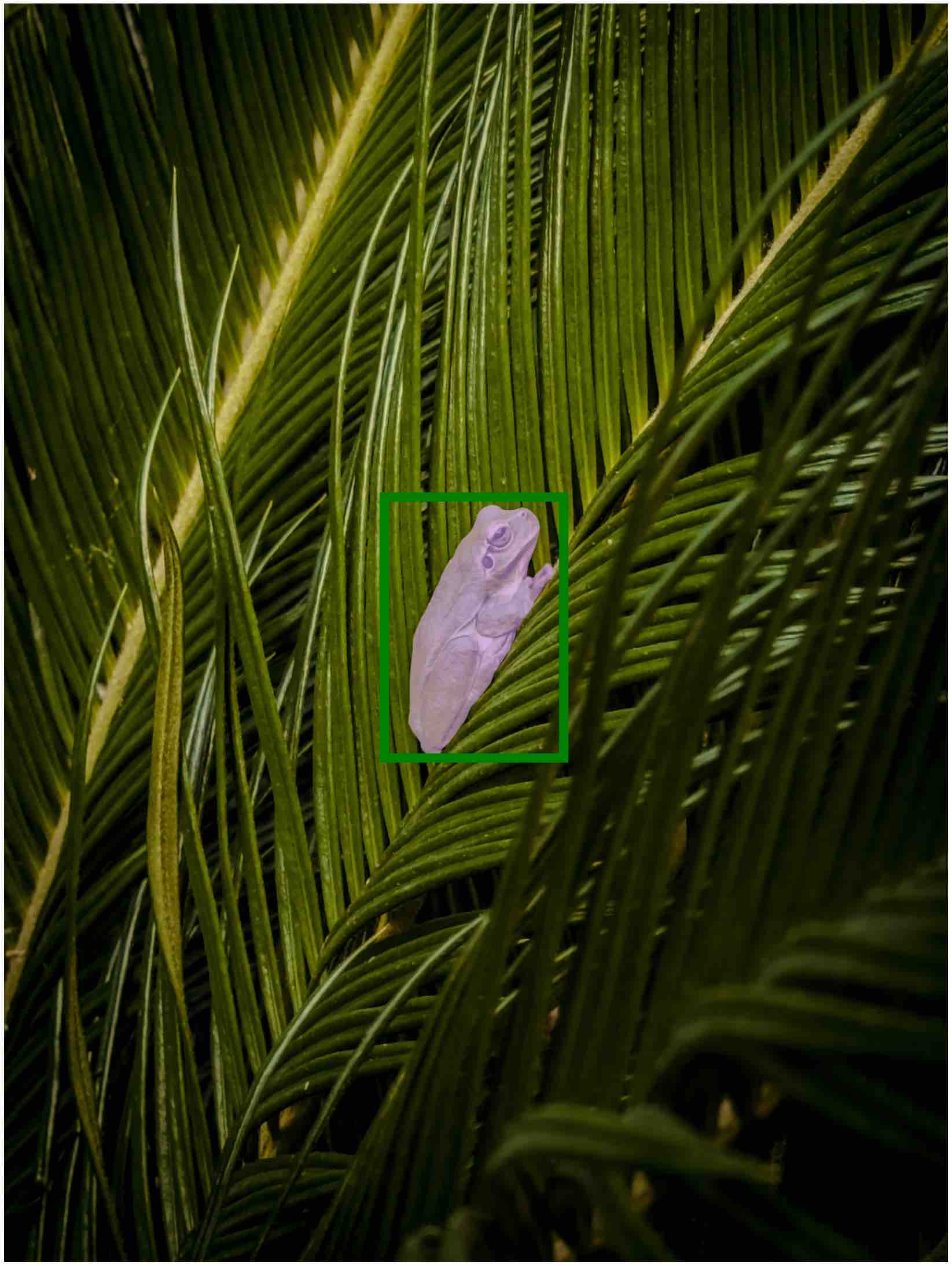}
        \caption{Processed image}
        \label{fig:frog_res}
    \end{minipage}   
\end{figure}
\\
\FloatBarrier
\noindent \textbf{Case 5:} Given the prompt "highlight all frogs by masking them" and an image:\\
\begin{figure}[htbp]
    \centering
    \begin{minipage}{0.45\linewidth}
        \includegraphics[width=\linewidth]{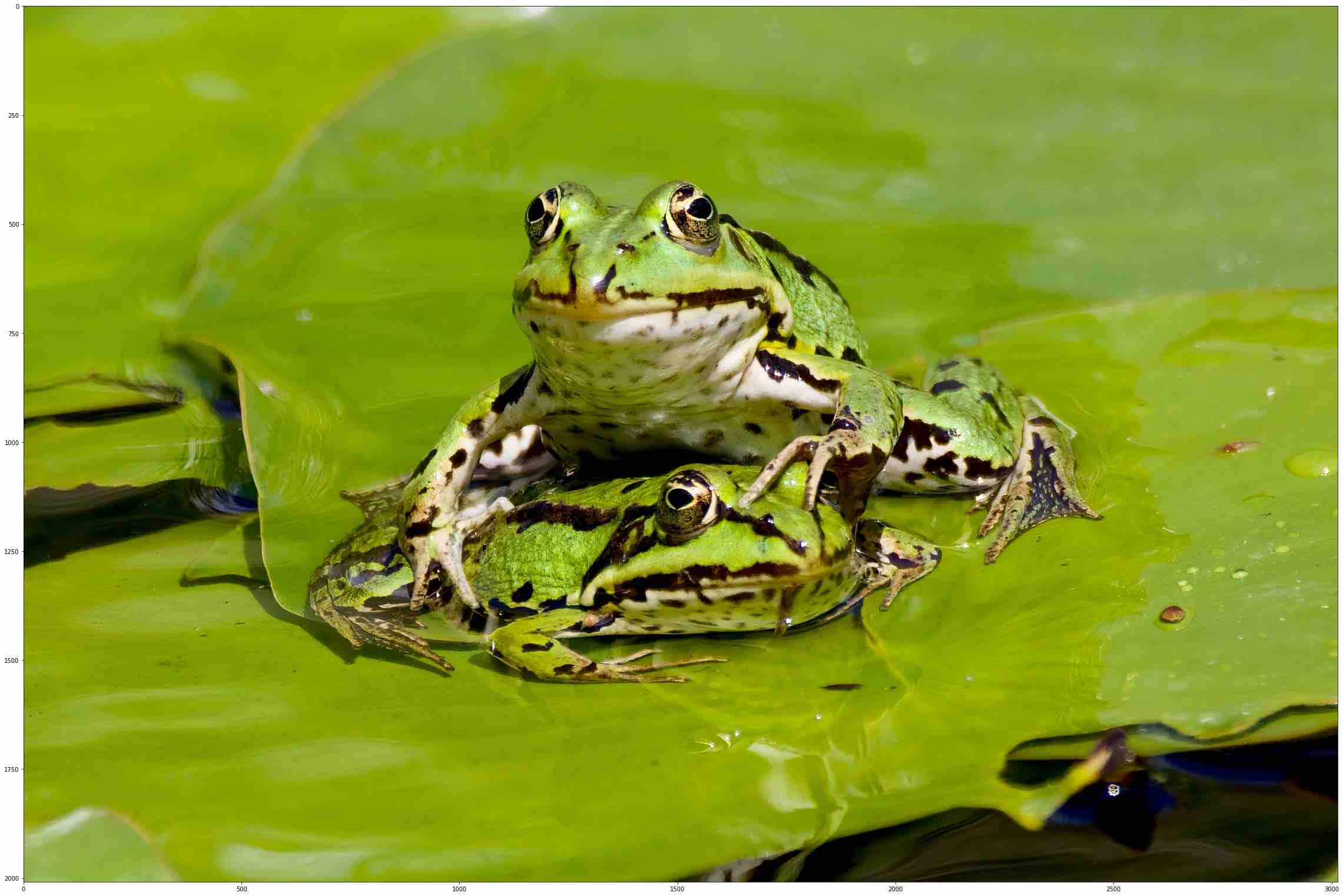}
        \caption{Original image}
        \label{fig:frogs_src}
    \end{minipage}
    \hfill
    \begin{minipage}{0.45\linewidth}
        \includegraphics[width=\linewidth]{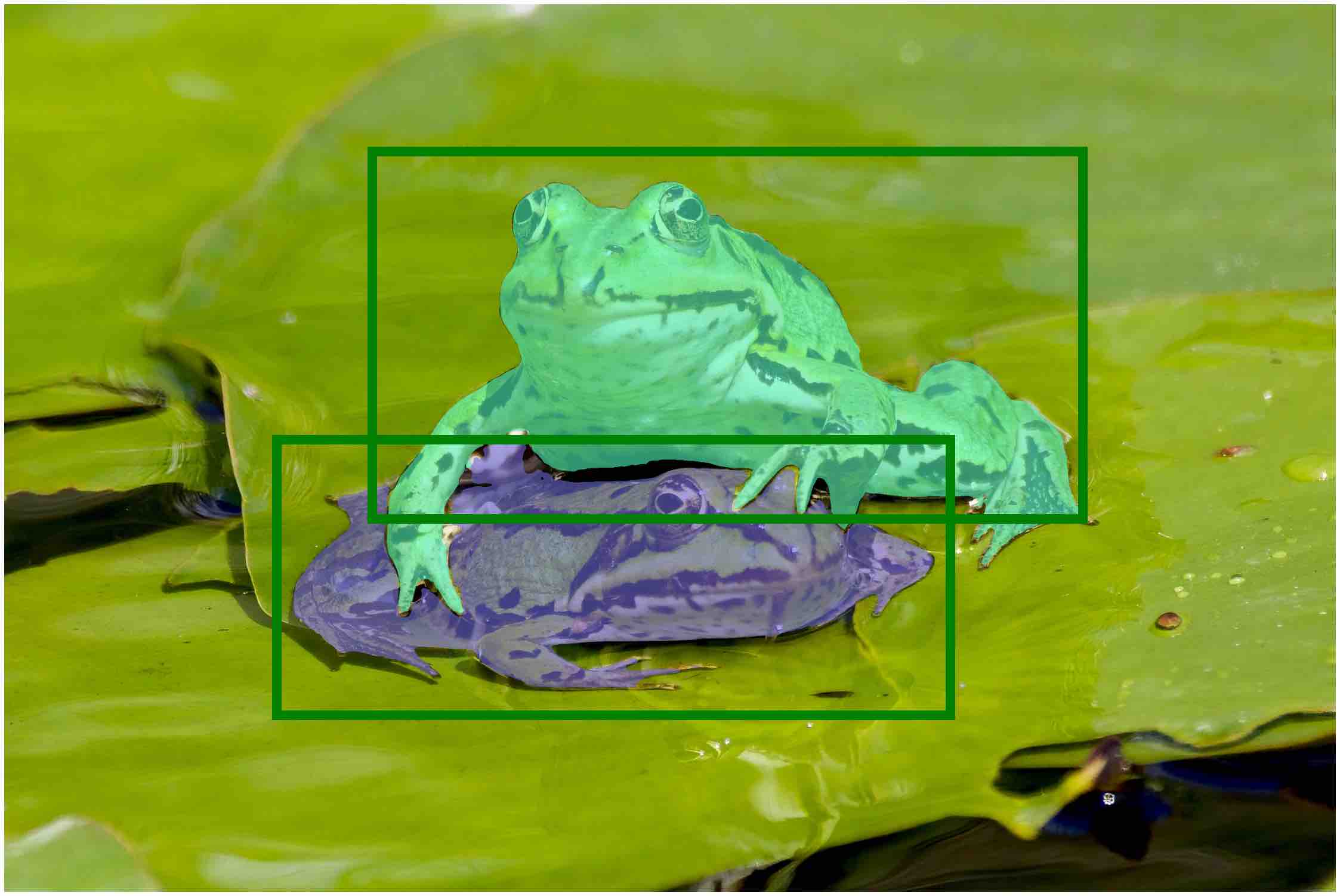}
        \caption{Processed image}
        \label{fig:frogs_res}
    \end{minipage}   
\end{figure}
\\
\FloatBarrier
\noindent \textbf{Case 6:} Given the prompt "mask out the main object in the image" and an image:\\
\begin{figure}[htbp]
    \centering
    \begin{minipage}{0.45\linewidth}
        \includegraphics[width=\linewidth]{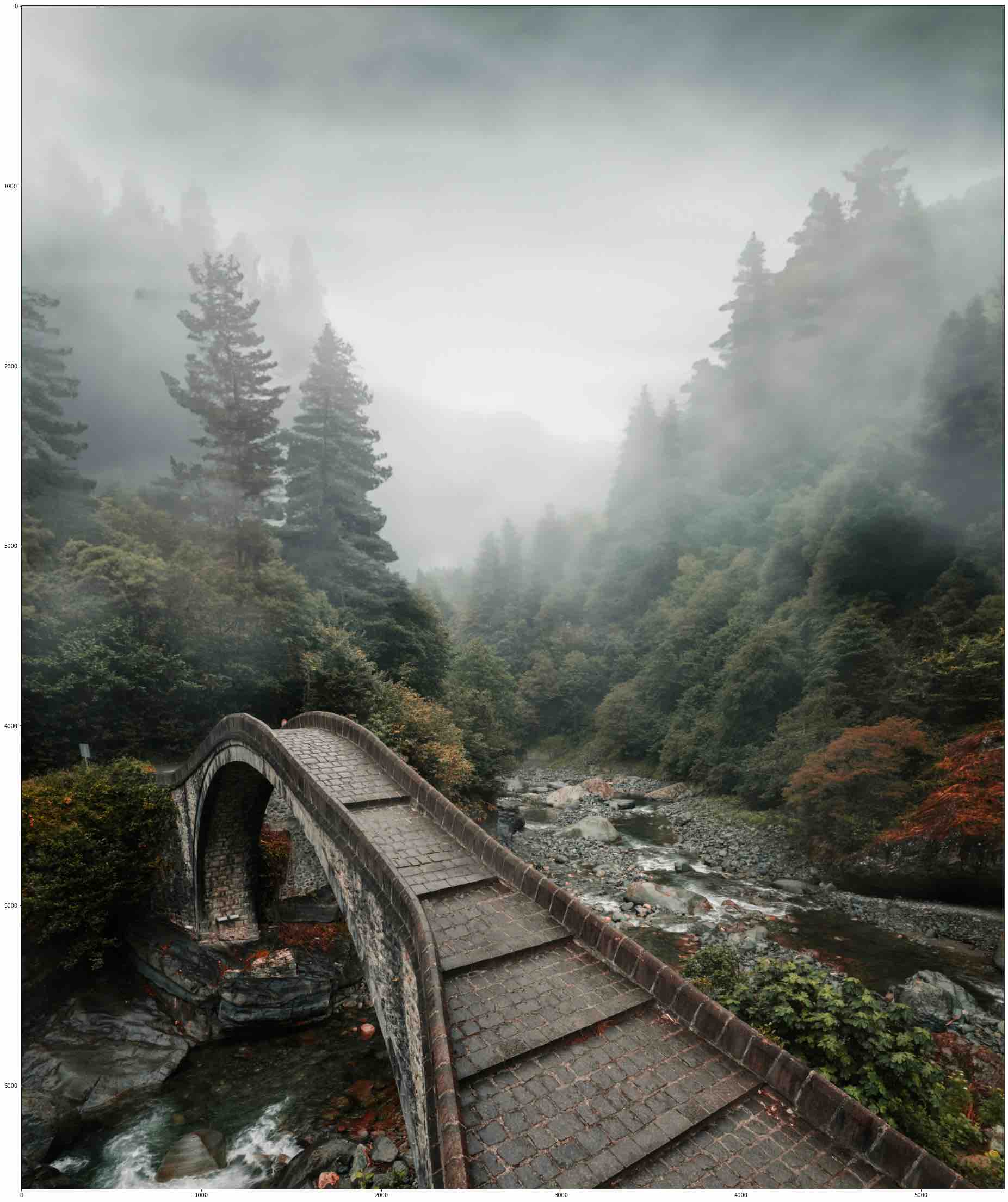}
        \caption{Original image}
        \label{fig:bridge_src}
    \end{minipage}
    \hfill
    \begin{minipage}{0.45\linewidth}
        \includegraphics[width=\linewidth]{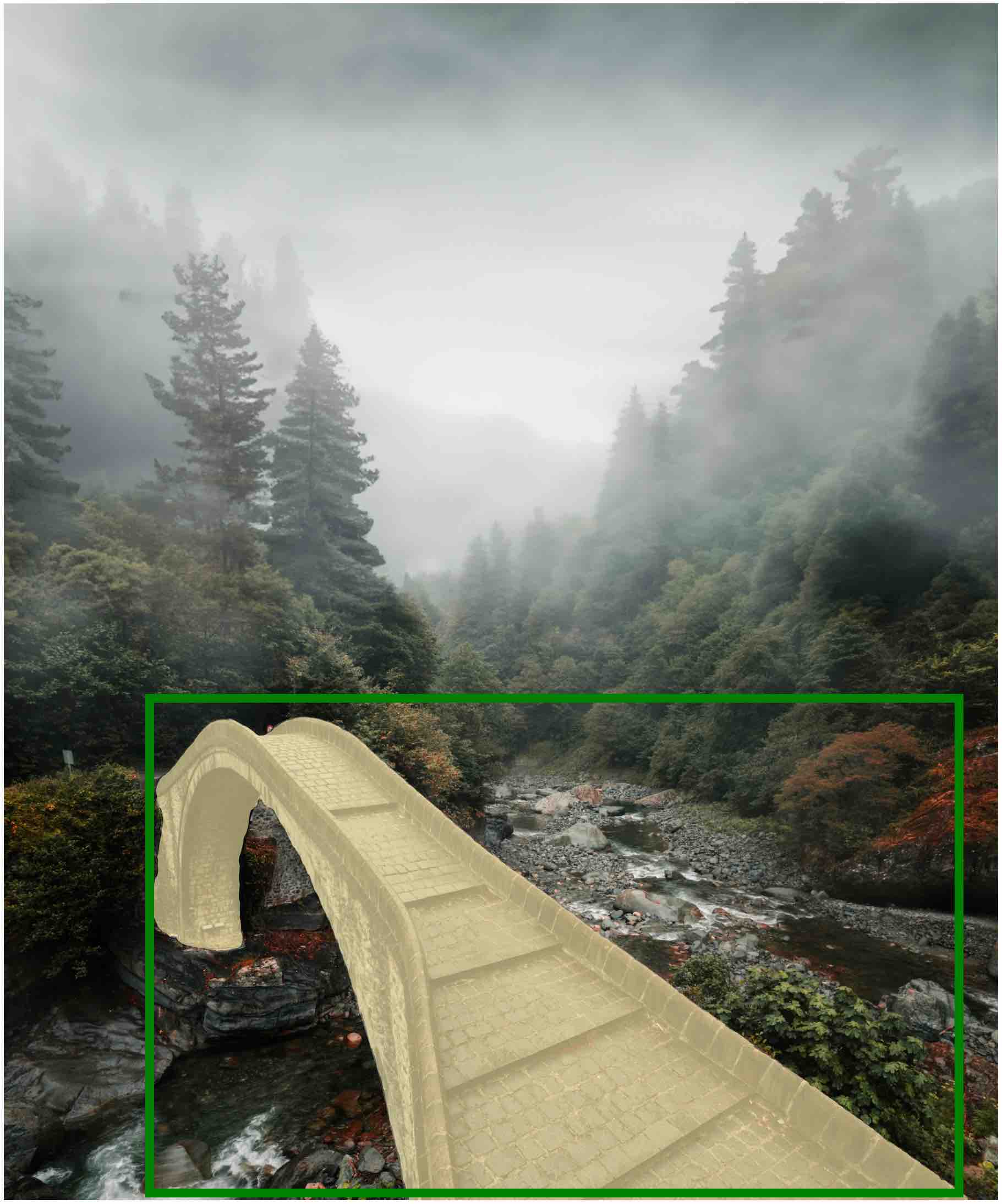}
        \caption{Processed image}
        \label{fig:bridge_res}
    \end{minipage}   
\end{figure}
\\
\newpage
\FloatBarrier
\noindent \textbf{Case 7:} Given the prompt "Can you see a bird? Please mask it if so." and an image:\\
\begin{figure}[htbp]
    \centering
    \begin{minipage}{0.45\linewidth}
        \includegraphics[width=\linewidth]{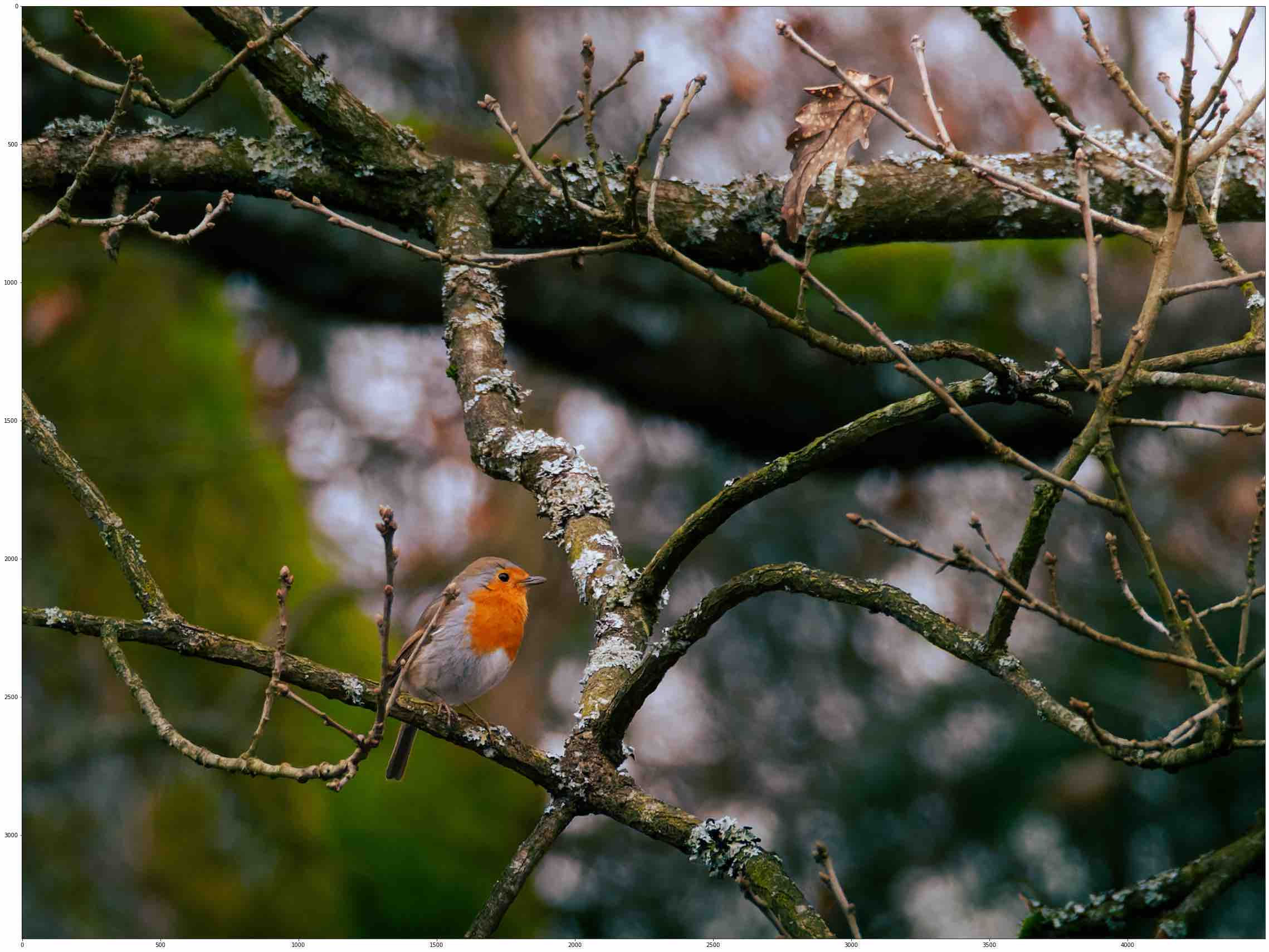}
        \caption{Original image}
        \label{fig:bird_src}
    \end{minipage}
    \hfill
    \begin{minipage}{0.45\linewidth}
        \includegraphics[width=\linewidth]{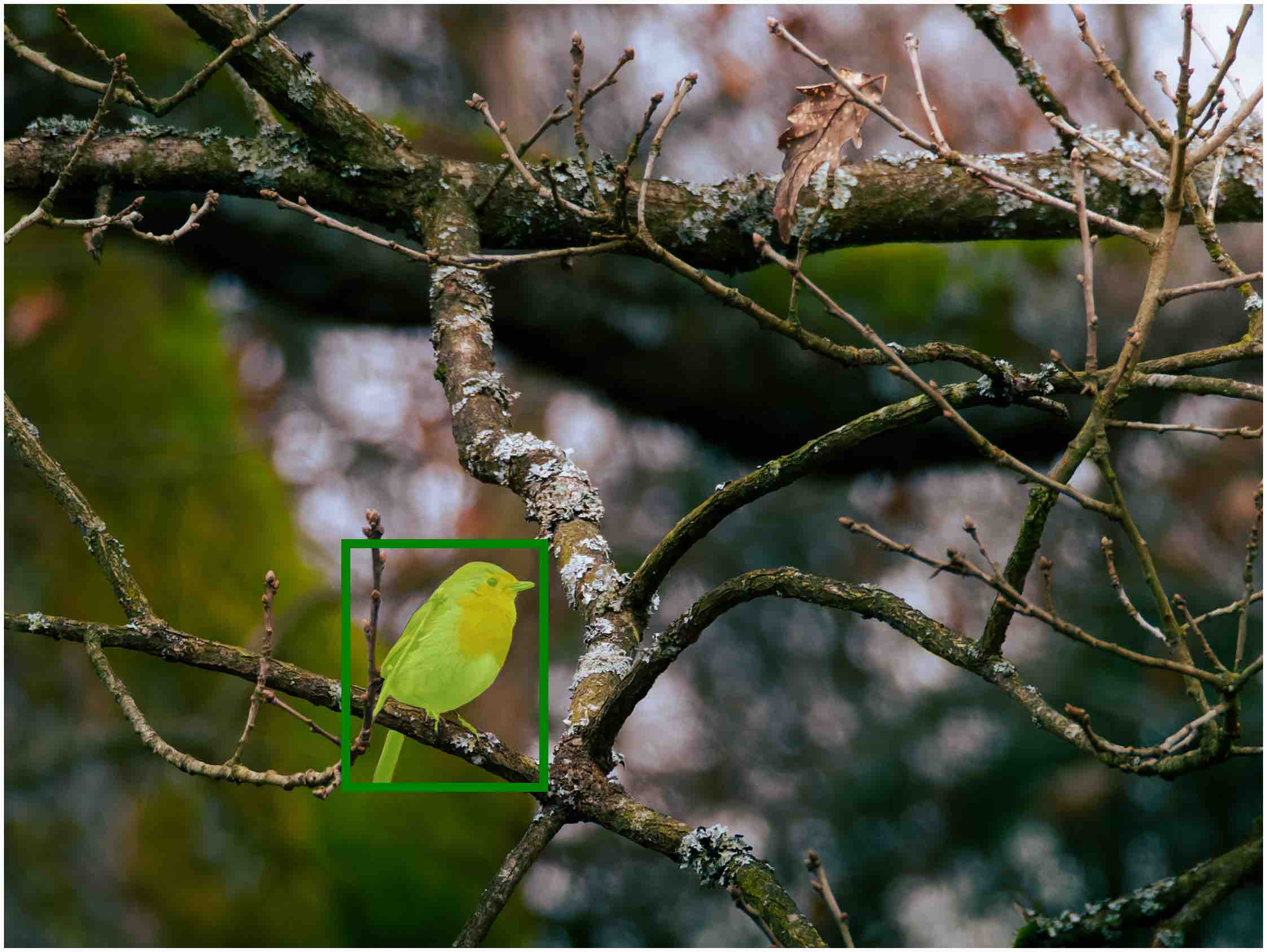}
        \caption{Processed image}
        \label{fig:bird_res}
    \end{minipage}   
\end{figure}
\\
\FloatBarrier
\noindent \textbf{Case 8:} Given the prompt "Detect and segment the bird using more than one foundation models." and an image:\\
\begin{figure}[htbp]
    \centering
    \begin{minipage}{0.45\linewidth}
        \includegraphics[width=\linewidth]{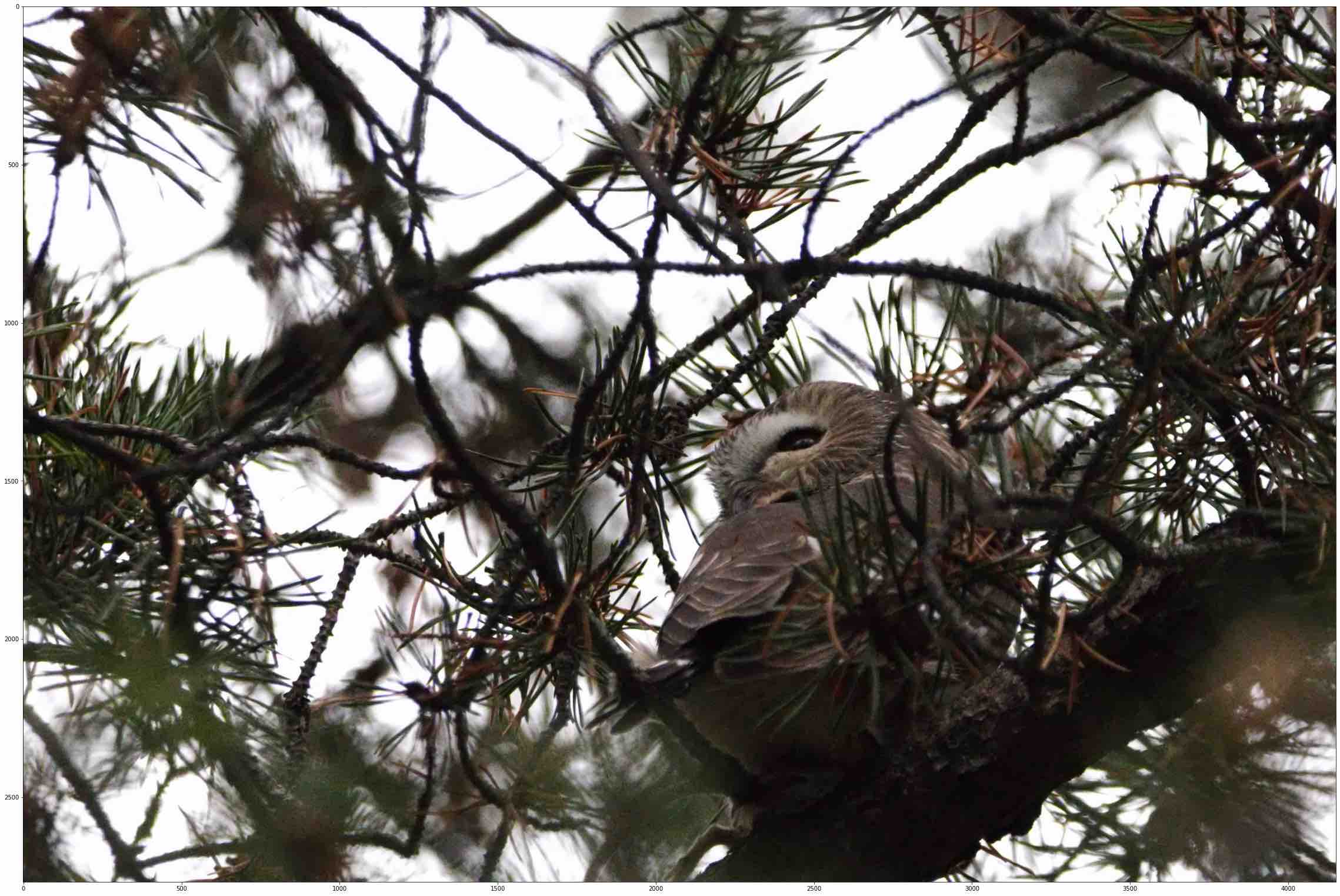}
        \caption{Original image}
        \label{fig:bd_src}
    \end{minipage}
    \hfill
    \begin{minipage}{0.45\linewidth}
        \includegraphics[width=\linewidth]{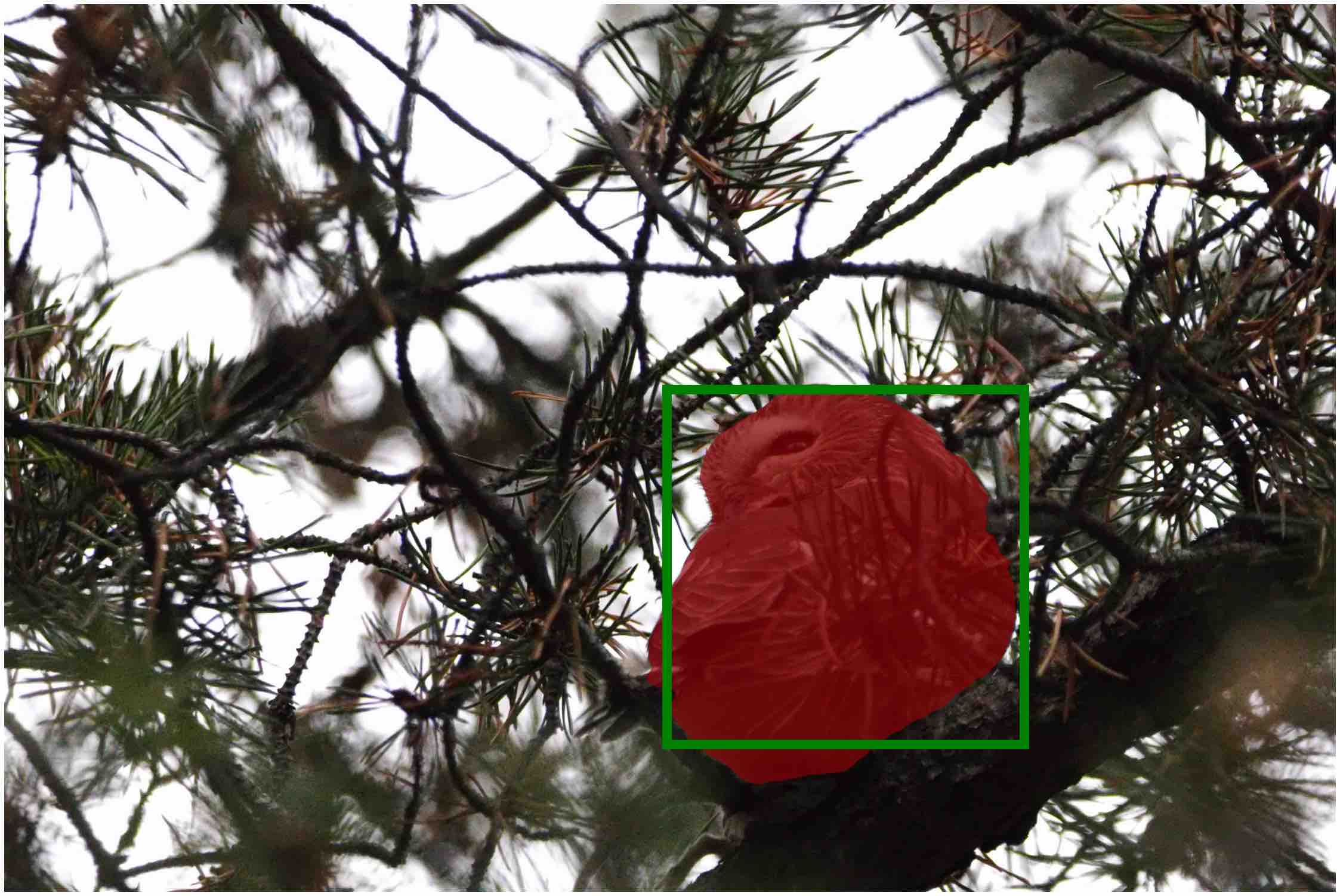}
        \caption{Processed image}
        \label{fig:bd_res}
    \end{minipage}   
\end{figure}
\\
\FloatBarrier
\noindent \textbf{Case 9:} Given the prompt "Mask any building in the image." and an image:\\
\begin{figure}[htbp]
    \centering
    \begin{minipage}{0.45\linewidth}
        \includegraphics[width=\linewidth]{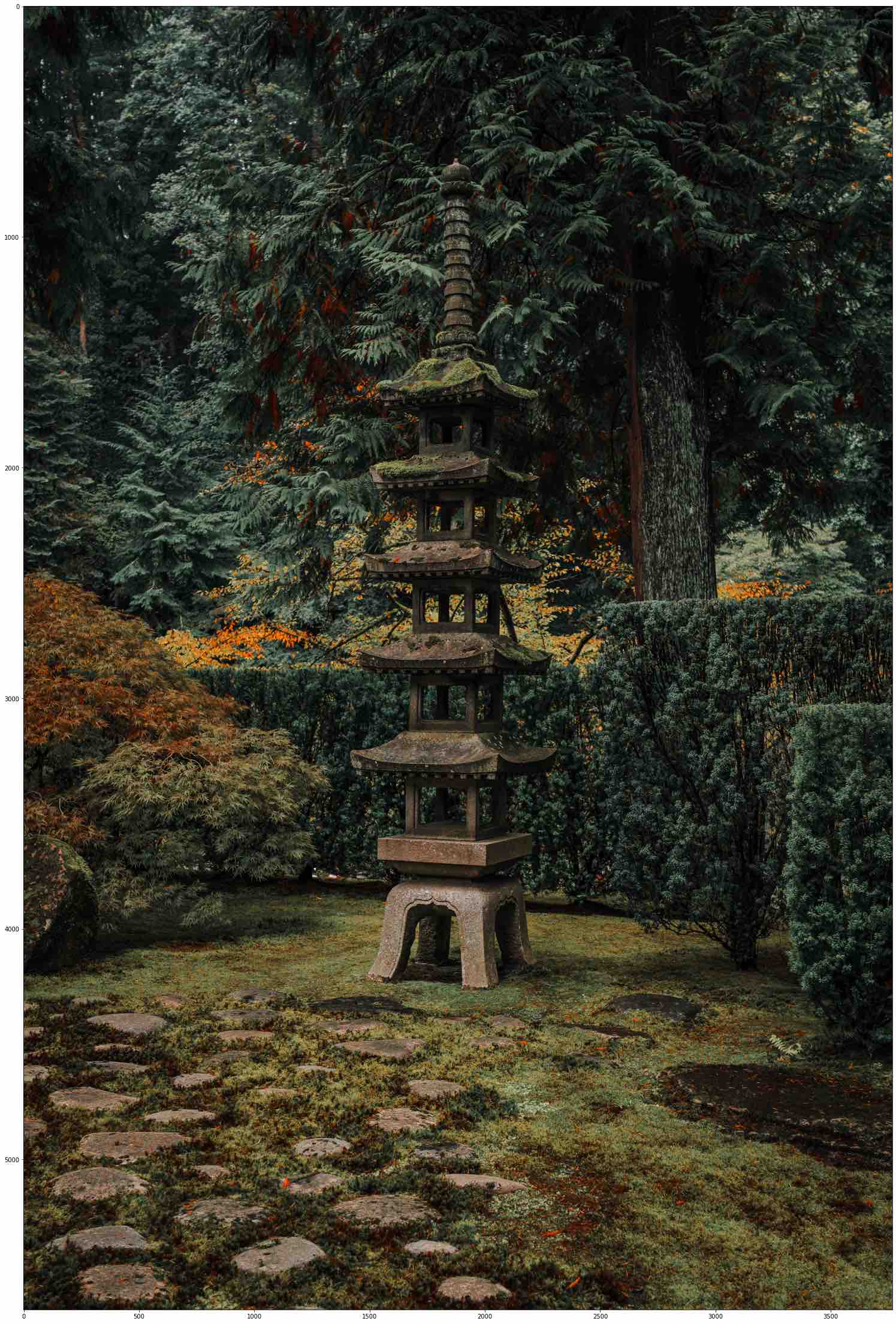}
        \caption{Original image}
        \label{fig:tower_src}
    \end{minipage}
    \hfill
    \begin{minipage}{0.45\linewidth}
        \includegraphics[width=\linewidth]{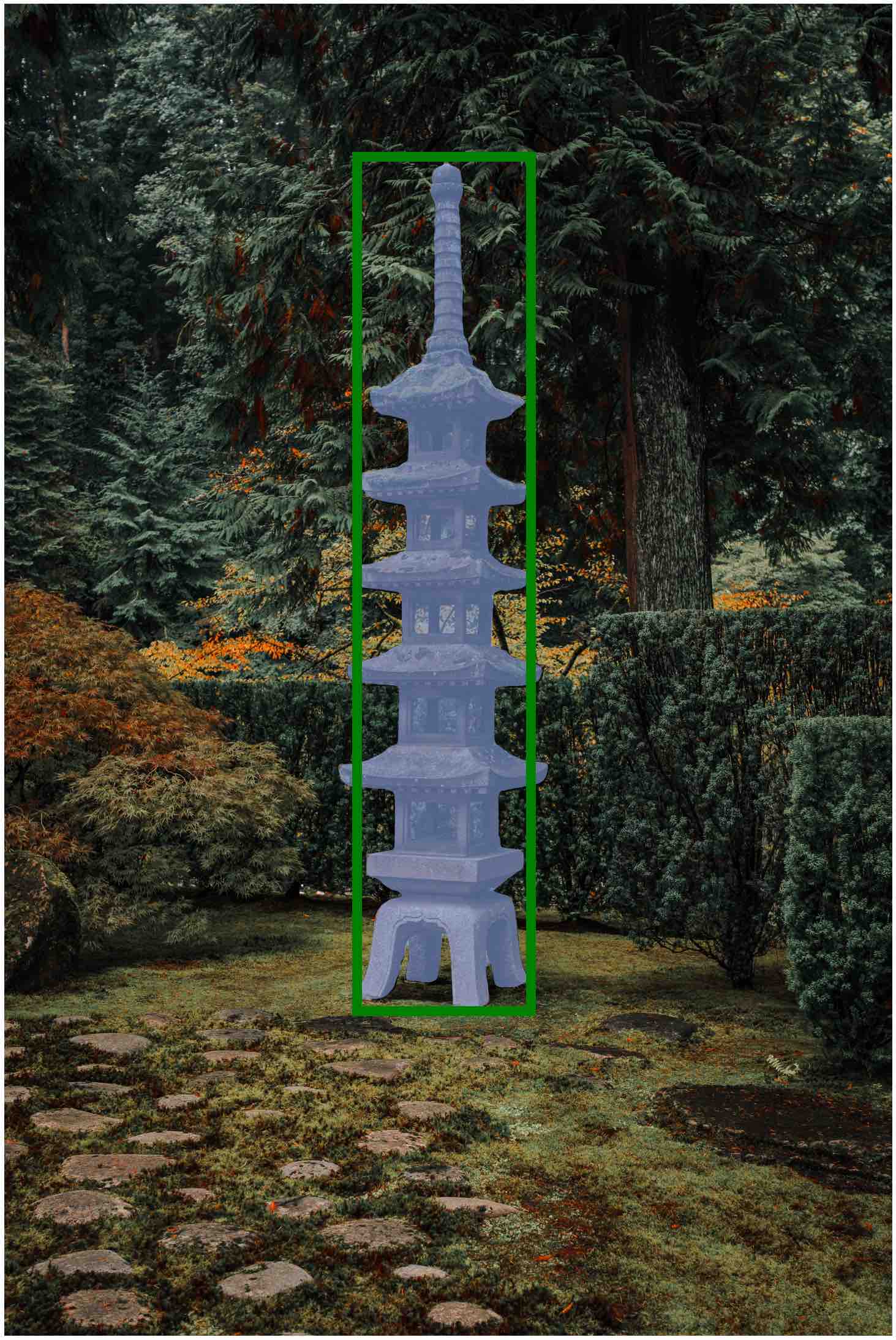}
        \caption{Processed image}
        \label{fig:tower_res}
    \end{minipage}   
\end{figure}
\\
\newpage
\FloatBarrier
\noindent Now let's find anomaly objects in the images without specifying an object name. Please see a few examples below:\\
\\
\noindent \textbf{Case 10:} Given the prompt "identify any anomaly object and segment it if have" and an image:\\
\begin{figure}[htbp]
    \centering
    \begin{minipage}{0.45\linewidth}
        \includegraphics[width=\linewidth]{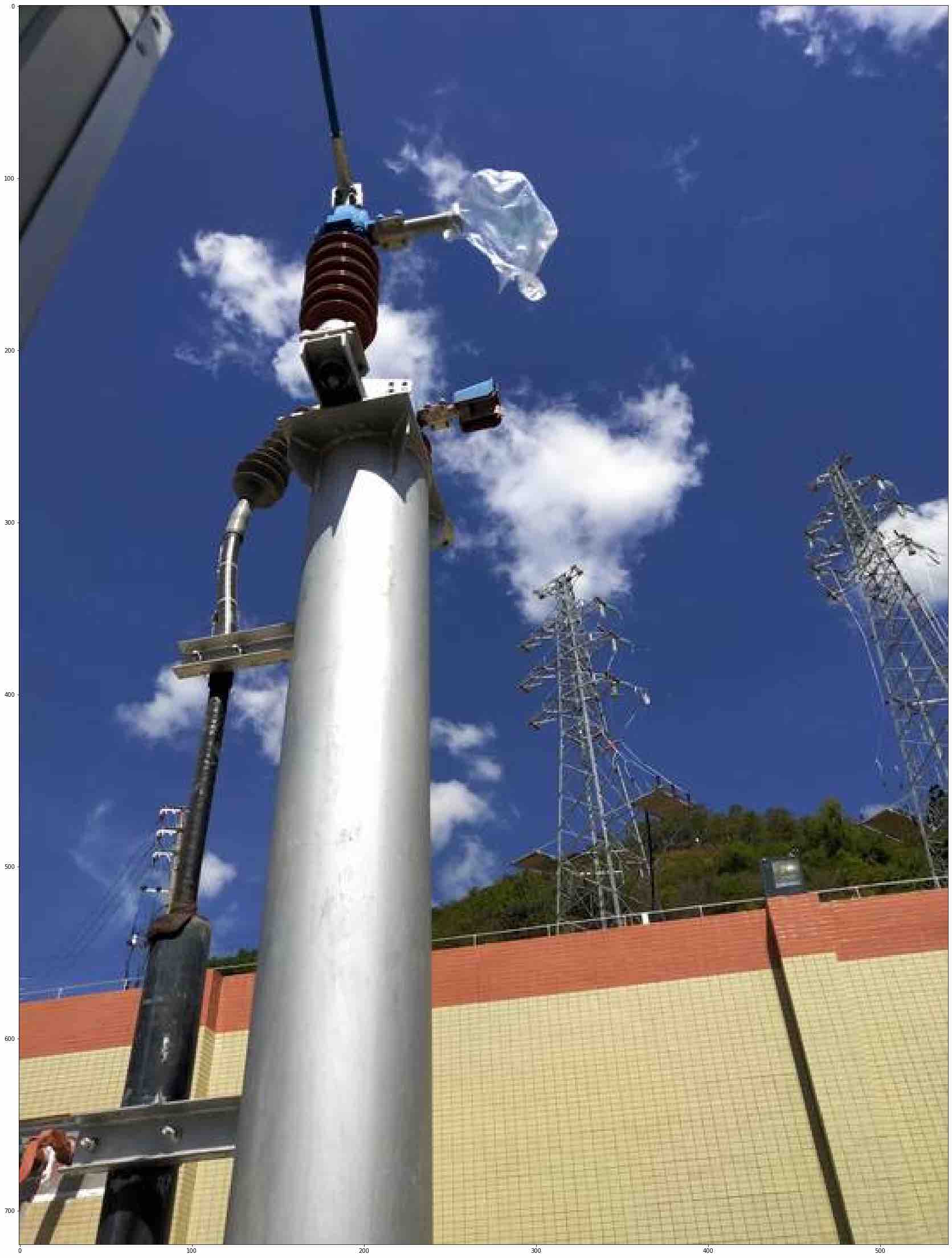}
        \caption{Original image}
        \label{fig:anomaly_src}
    \end{minipage}
    \hfill
    \begin{minipage}{0.45\linewidth}
        \includegraphics[width=\linewidth]{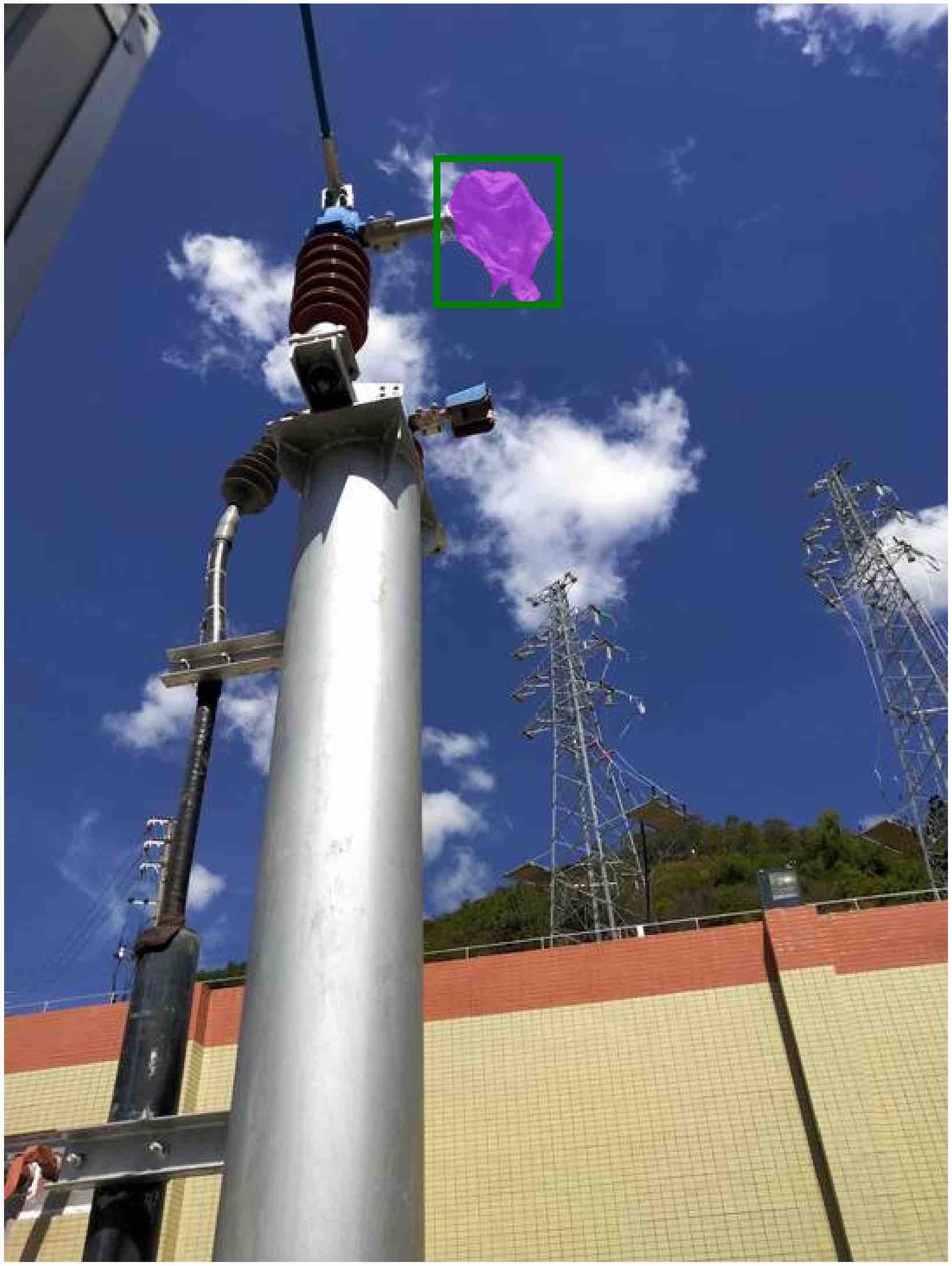}
        \caption{Processed image}
        \label{fig:anomaly_res}
    \end{minipage}   
\end{figure}
\\
\FloatBarrier
\noindent \textbf{Case 11:} Given the prompt "find any anomaly object and detect/segment it" and an image:\\
\begin{figure}[htbp]
    \centering
    \begin{minipage}{0.45\linewidth}
        \includegraphics[width=\linewidth]{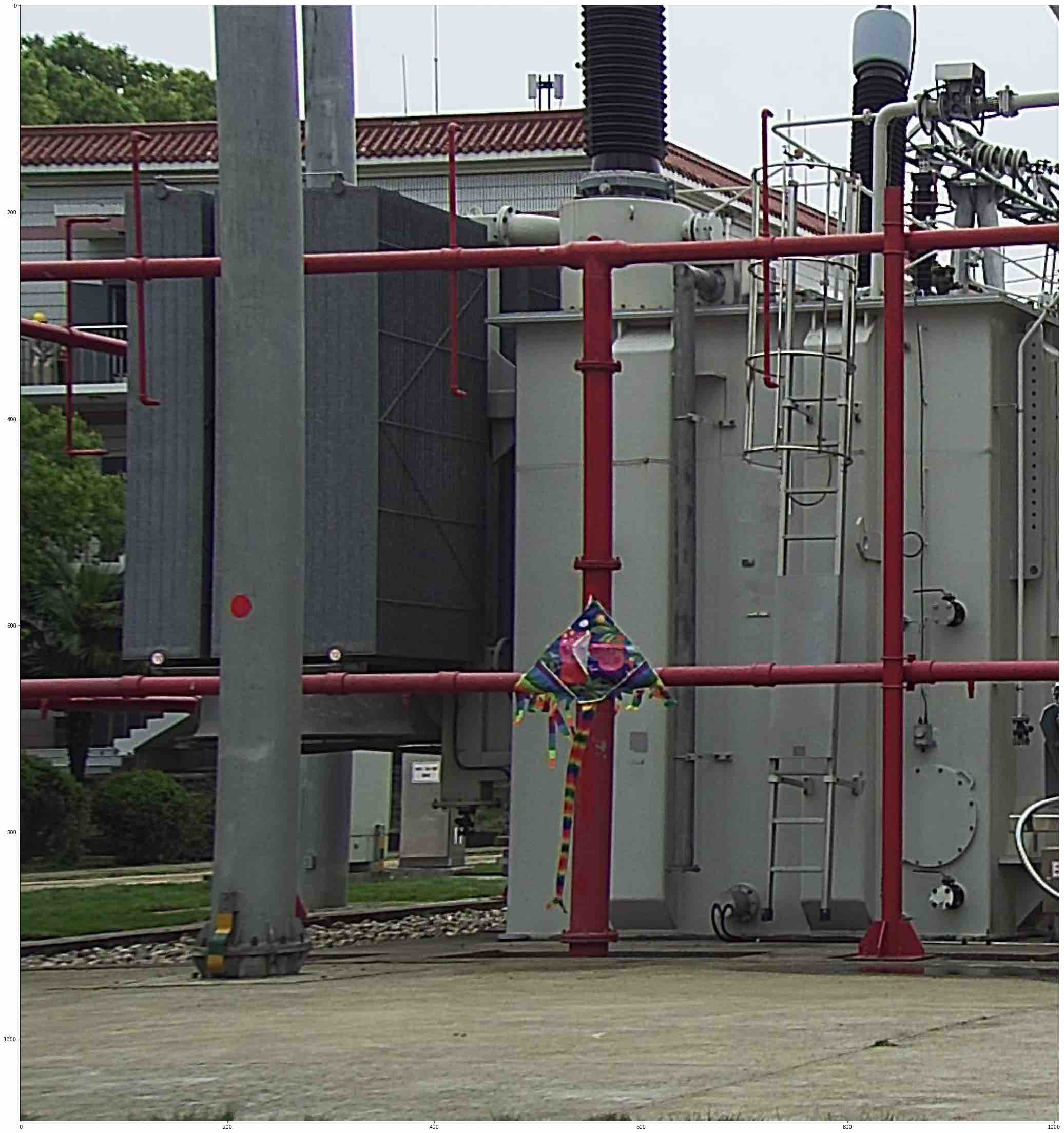}
        \caption{Original image}
        \label{fig:ano_src}
    \end{minipage}
    \hfill
    \begin{minipage}{0.45\linewidth}
        \includegraphics[width=\linewidth]{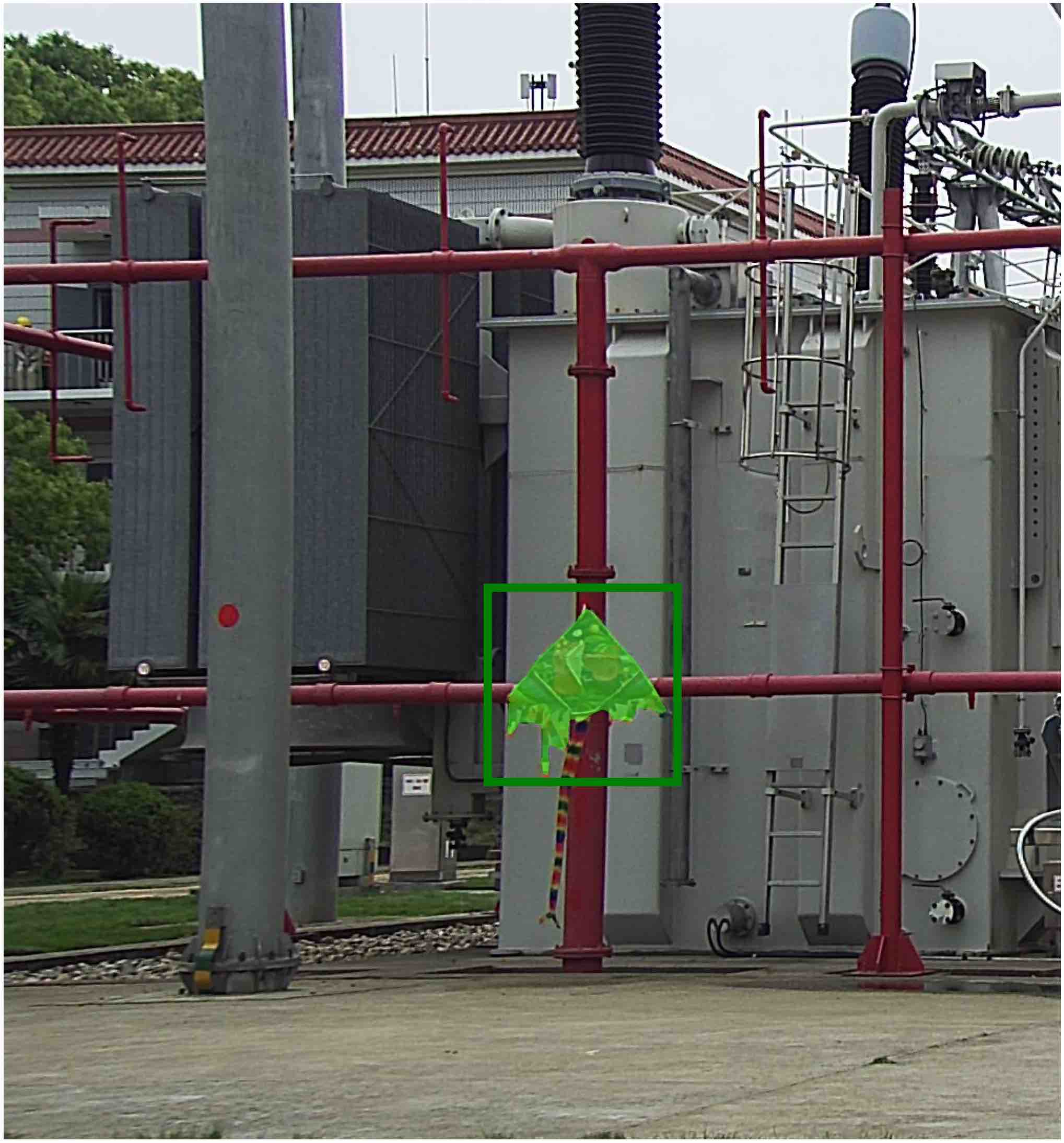}
        \caption{Processed image}
        \label{fig:ano_res}
    \end{minipage}   
\end{figure}
\\
\FloatBarrier
\noindent \textbf{Case 12:} Given the prompt "find a different animal and segment it" and an image:\\
\begin{figure}[htbp]
    \centering
    \begin{minipage}{0.45\linewidth}
        \includegraphics[width=\linewidth]{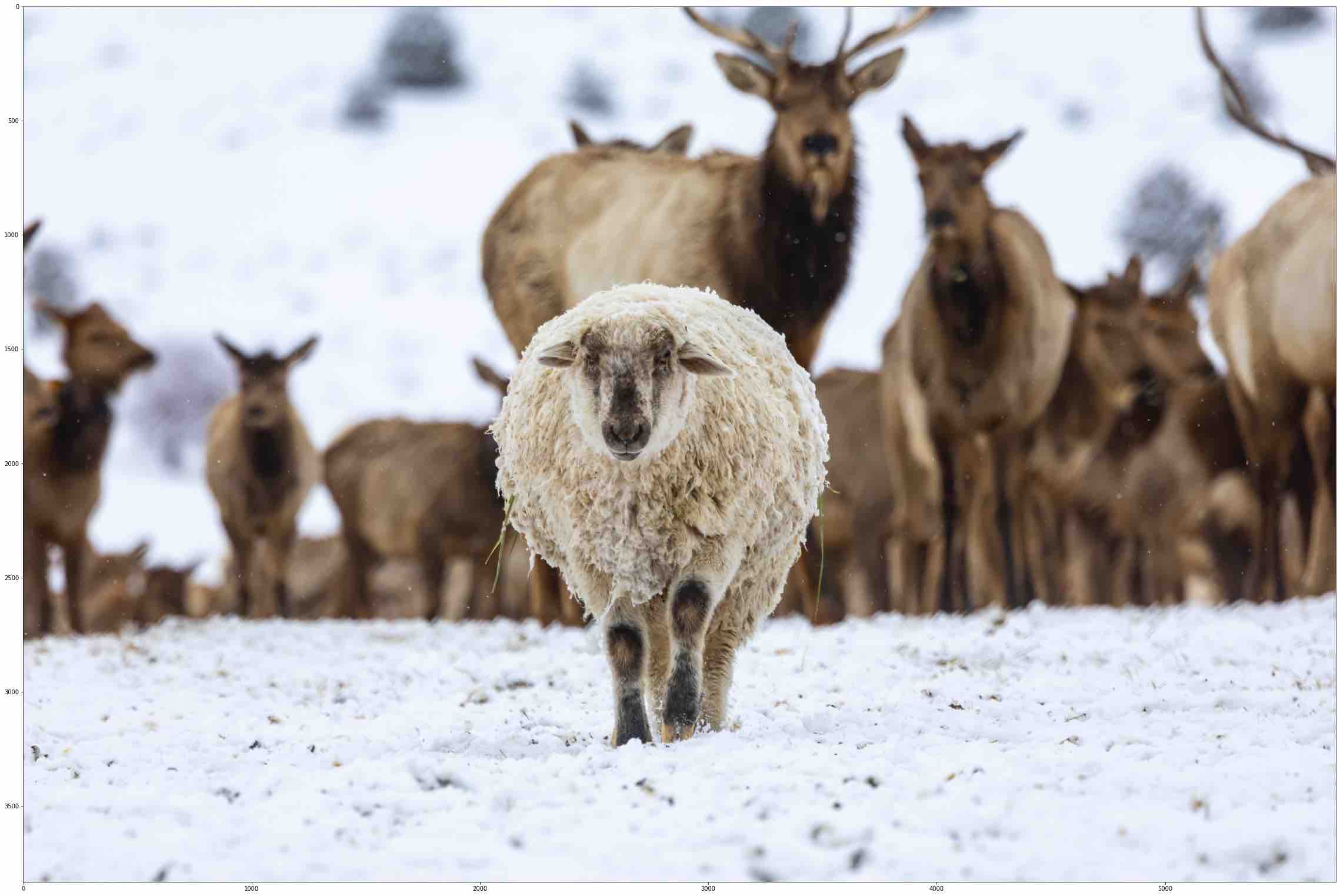}
        \caption{Original image}
        \label{fig:sheep_src}
    \end{minipage}
    \hfill
    \begin{minipage}{0.45\linewidth}
        \includegraphics[width=\linewidth]{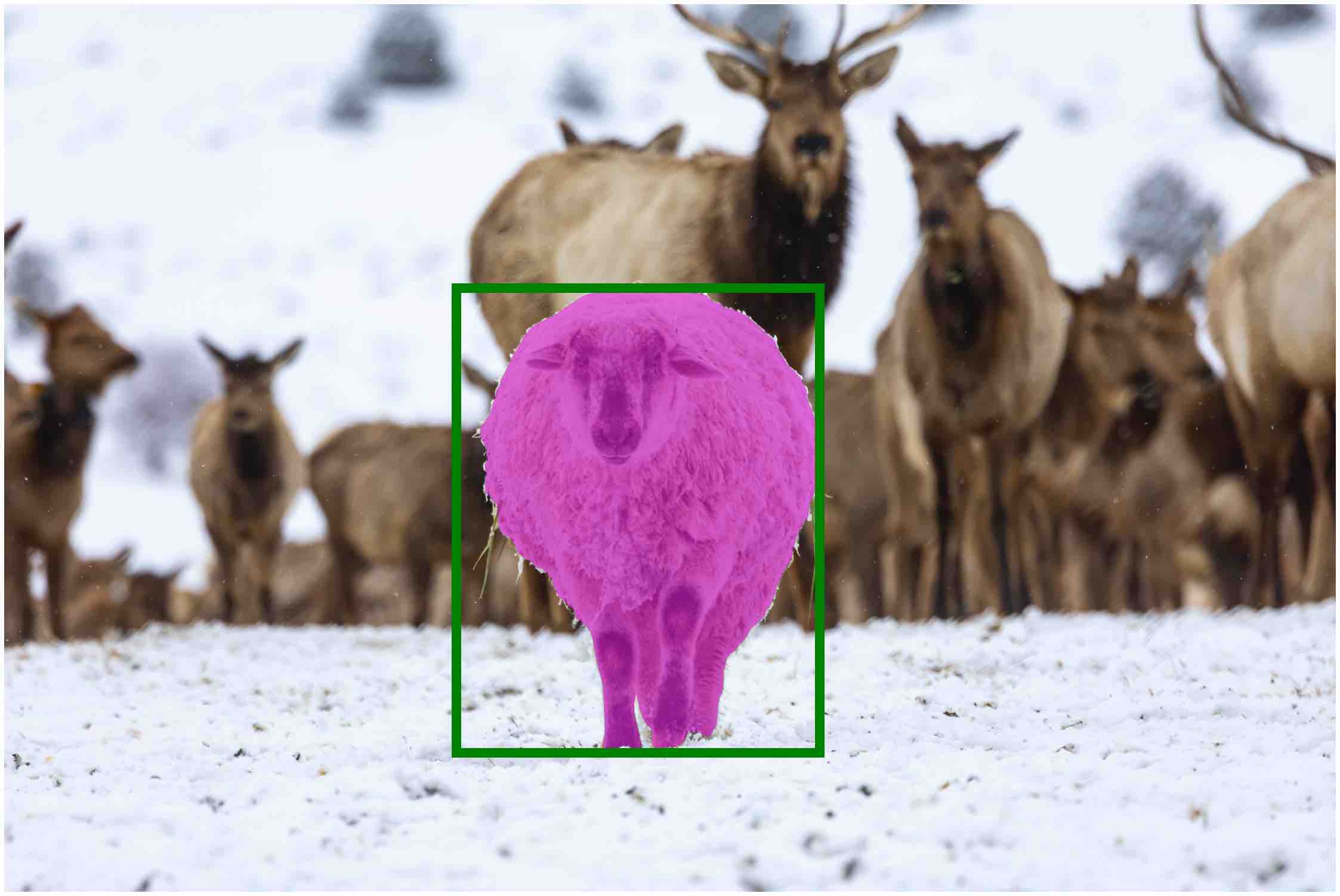}
        \caption{Processed image}
        \label{fig:sheep_res}
    \end{minipage}   
\end{figure}
\\
\section{Limitations}

While UnifiedVisionGPT presents a substantial advancement in integrating state-of-the-art (SOTA) computer vision models with large language models (LLMs), it is not without its limitations. A primary constraint lies in the rapid evolution of SOTA vision foundation models and vision-language models, which poses a challenge to maintaining the relevance and effectiveness of UnifiedVisionGPT. The continual emergence of new models and techniques necessitates frequent updates and adaptations to the UnifiedVisionGPT framework to ensure compatibility and optimal performance.\\
\\
Another limitation stems from UnifiedVisionGPT's reliance on the integration of multiple expert models. While this design enhances versatility and ensures the utilization of the best features from various models, it also introduces complexity in model management and coordination. In order to ensure seamless interaction and data flow between frequently evolving models, meticulous design and maintenance is required.\\
\\
Additionally, while UnifiedVisionGPT aims to act as a vision-language model, its performance is contingent on the quality and capabilities of the integrated vision models and LLMs. Any shortcomings or biases in these underlying models could potentially propagate through the UnifiedVisionGPT framework, affecting the overall performance and reliability of the system.\\
\\
In conclusion, although UnifiedVisionGPT represents a significant stride toward a unified and versatile AI system, it is not without its limitations. Ensuring the seamless integration and management of various expert models while also facing the challenges posed by the fast-paced advancements in vision foundation models highlight critical areas for future work and improvement.\\
\\
\section{Conclusion}

In this research paper, we have presented UnifiedVisionGPT, a novel framework that integrates state-of-the-art computer vision models with large language models (LLMs). This framework offers a robust and adaptable platform for object detection and various AI tasks. UnifiedVisionGPT leverages the synergistic capabilities of expert models, such as YOLO and SAM, and LLMs to provide a seamless user experience, automating the entire workflow from vision pre-processing to post-processing.\\
\\
Our framework distinguishes itself by not only executing vision-oriented tasks but also understanding and interpreting user requests through natural language processing. The integration of LLMs enables UnifiedVisionGPT to extract context, details, and intent from user inputs, translating them into precise object analysis and recognition. This results in a highly adaptive system that can tailor its responses to the nuances of user language and the specific content of visual data.\\
\\
Through our discussions in the paper, including the related works and the intricacies of the UnifiedVisionGPT framework, we have demonstrated the unique position and capabilities of UnifiedVisionGPT in the current landscape of AI and computer vision. UnifiedVisionGPT sets a new standard for multimodal AI applications, ensuring efficiency, versatility, and performance.\\
\\
As we look to the future, the potential for UnifiedVisionGPT to evolve and integrate with upcoming LLMs and vision models is vast, promising even more personalized and context-aware interactions. This research lays the groundwork for future developments in this domain, aiming to continually enhance and tailor AI systems to meet the diverse and growing needs of users.\\
\\
{
    \small
    \bibliographystyle{ieeenat_fullname}
    \bibliography{main}

\begin{thebibliography}{26}
\providecommand{\natexlab}[1]{#1}
\providecommand{\url}[1]{\texttt{#1}}
\expandafter\ifx\csname urlstyle\endcsname\relax
  \providecommand{\doi}[1]{doi: #1}\else
  \providecommand{\doi}{doi: \begingroup \urlstyle{rm}\Url}\fi

\bibitem[Alayrac et~al.(2022)Alayrac, Donahue, Luc, Miech, Barr, Hasson, Lenc, Mensch, Millican, Reynolds, Ring, Rutherford, Cabi, Han, Gong, Samangooei, Monteiro, Menick, Borgeaud, Brock, Nematzadeh, Sharifzadeh, Binkowski, Barreira, Vinyals, Zisserman, and Simonyan]{flamingo}
Jean-Baptiste Alayrac, Jeff Donahue, Pauline Luc, Antoine Miech, Iain Barr, Yana Hasson, Karel Lenc, Arthur Mensch, Katie Millican, Malcolm Reynolds, Roman Ring, Eliza Rutherford, Serkan Cabi, Tengda Han, Zhitao Gong, Sina Samangooei, Marianne Monteiro, Jacob Menick, Sebastian Borgeaud, Andrew Brock, Aida Nematzadeh, Sahand Sharifzadeh, Mikolaj Binkowski, Ricardo Barreira, Oriol Vinyals, Andrew Zisserman, and Karen Simonyan.
\newblock Flamingo: a visual language model for few-shot learning, 2022.

\bibitem[Bao et~al.(2022)Bao, Wang, Dong, Liu, Mohammed, Aggarwal, Som, and Wei]{vlmo}
Hangbo Bao, Wenhui Wang, Li Dong, Qiang Liu, Owais~Khan Mohammed, Kriti Aggarwal, Subhojit Som, and Furu Wei.
\newblock Vlmo: Unified vision-language pre-training with mixture-of-modality-experts, 2022.

\bibitem[Betker et~al.(2023)Betker, Goh, Jing, Brooks, Wang, Li, Ouyang, Zhuang, Lee, Guo, Manassra, Dhariwal, Chu, Jiao, and Ramesh]{dalle}
James Betker, Gabriel Goh, Li Jing, Tim Brooks, Jianfeng Wang, Linjie Li, Long Ouyang, Juntang Zhuang, Joyce Lee, Yufei Guo, Wesam Manassra, Prafulla Dhariwal, Casey Chu, Yunxin Jiao, and Aditya Ramesh.
\newblock Improving image generation with better captions.
\newblock https://cdn.openai.com/papers/dall-e-3.pdf, 2023.

\bibitem[Chen et~al.(2023{\natexlab{a}})Chen, Wang, and Yu]{gsam}
Yuxin Chen, Jingdong Wang, and Fisher Yu.
\newblock Grounded segment anything model: A neural architecture for self-supervised instance segmentation with grounded language instructions, 2023{\natexlab{a}}.

\bibitem[Chen et~al.(2023{\natexlab{b}})Chen, Wang, and Yu]{sam}
Yuxin Chen, Jingdong Wang, and Fisher Yu.
\newblock Scaling factors for efficient neural architecture search, 2023{\natexlab{b}}.

\bibitem[Darcet et~al.(2023)Darcet, Oquab, Mairal, and Bojanowski]{darcet}
Timothée Darcet, Maxime Oquab, Julien Mairal, and Piotr Bojanowski.
\newblock Vision transformers need registers, 2023.

\bibitem[Face(2023)]{huggingface}
Hugging Face.
\newblock Hugging face: An open-source community for machine learning, 2023.

\bibitem[Fu et~al.(2022)Fu, Li, Gan, Lin, Wang, Wang, and Liu]{violet}
Tsu-Jui Fu, Linjie Li, Zhe Gan, Kevin Lin, William~Yang Wang, Lijuan Wang, and Zicheng Liu.
\newblock Violet : End-to-end video-language transformers with masked visual-token modeling, 2022.

\bibitem[Gao et~al.(2023)Gao, Sarkar, Xia, Xiao, Wu, Ichter, Majumdar, and Sadigh]{pgvlm}
Jensen Gao, Bidipta Sarkar, Fei Xia, Ted Xiao, Jiajun Wu, Brian Ichter, Anirudha Majumdar, and Dorsa Sadigh.
\newblock Physically grounded vision-language models for robotic manipulation, 2023.

\bibitem[Jocher et~al.(2023)Jocher, Chaurasia, and Qiu]{yolo}
G. Jocher, A. Chaurasia, and J. Qiu.
\newblock Yolov8 by ultralytics.
\newblock https://github.com/ultralytics/ultralytics, 2023.

\bibitem[Kojima et~al.(2023)Kojima, Gu, Reid, Matsuo, and Iwasawa]{llmzs}
Takeshi Kojima, Shixiang~Shane Gu, Machel Reid, Yutaka Matsuo, and Yusuke Iwasawa.
\newblock Large language models are zero-shot reasoners, 2023.

\bibitem[OpenAI(2023)]{gpt4}
OpenAI.
\newblock Gpt-4 technical report.
\newblock Technical report, OpenAI, 2023.

\bibitem[Oquab et~al.(2023)Oquab, Darcet, Moutakanni, Vo, Szafraniec, Khalidov, Fernandez, Haziza, Massa, El-Nouby, Assran, Ballas, Galuba, Howes, Huang, Li, Misra, Rabbat, Sharma, Synnaeve, Xu, Jegou, Mairal, Labatut, Joulin, and Bojanowski]{dinov2}
Maxime Oquab, Timothée Darcet, Théo Moutakanni, Huy Vo, Marc Szafraniec, Vasil Khalidov, Pierre Fernandez, Daniel Haziza, Francisco Massa, Alaaeldin El-Nouby, Mahmoud Assran, Nicolas Ballas, Wojciech Galuba, Russell Howes, Po-Yao Huang, Shang-Wen Li, Ishan Misra, Michael Rabbat, Vasu Sharma, Gabriel Synnaeve, Hu Xu, Hervé Jegou, Julien Mairal, Patrick Labatut, Armand Joulin, and Piotr Bojanowski.
\newblock Dinov2: Learning robust visual features without supervision, 2023.

\bibitem[Radford et~al.(2021)Radford, Kim, Hallacy, Ramesh, Goh, Agarwal, Sastry, Askell, Mishkin, Clark, Krueger, and Sutskever]{clip}
Alec Radford, Jong~Wook Kim, Chris Hallacy, Aditya Ramesh, Gabriel Goh, Sandhini Agarwal, Girish Sastry, Amanda Askell, Pamela Mishkin, Jack Clark, Gretchen Krueger, and Ilya Sutskever.
\newblock Learning transferable visual models from natural language supervision, 2021.

\bibitem[Ramesh et~al.(2021)Ramesh, Pavlov, Goh, Gray, Voss, Radford, Chen, and Sutskever]{zeroshot}
Aditya Ramesh, Mikhail Pavlov, Gabriel Goh, Scott Gray, Chelsea Voss, Alec Radford, Mark Chen, and Ilya Sutskever.
\newblock Zero-shot text-to-image generation, 2021.

\bibitem[Shen et~al.(2023)Shen, Song, Tan, Li, Lu, and Zhuang]{hugginggpt}
Yongliang Shen, Kaitao Song, Xu Tan, Dongsheng Li, Weiming Lu, and Yueting Zhuang.
\newblock Hugginggpt: Solving ai tasks with chatgpt and its friends in hugging face, 2023.

\bibitem[Terven and Cordova-Esparza(2023)]{yoloreview}
Juan Terven and Diana Cordova-Esparza.
\newblock A comprehensive review of yolo: From yolov1 and beyond, 2023.

\bibitem[Touvron et~al.(2023{\natexlab{a}})Touvron, Lavril, Izacard, Martinet, Lachaux, Lacroix, Rozière, Goyal, Hambro, Azhar, Rodriguez, Joulin, Grave, and Lample]{llama}
Hugo Touvron, Thibaut Lavril, Gautier Izacard, Xavier Martinet, Marie-Anne Lachaux, Timothée Lacroix, Baptiste Rozière, Naman Goyal, Eric Hambro, Faisal Azhar, Aurelien Rodriguez, Armand Joulin, Edouard Grave, and Guillaume Lample.
\newblock Llama: Open and efficient foundation language models, 2023{\natexlab{a}}.

\bibitem[Touvron et~al.(2023{\natexlab{b}})Touvron, Martin, Stone, Albert, Almahairi, Babaei, Bashlykov, Batra, Bhargava, Bhosale, Bikel, Blecher, Ferrer, Chen, Cucurull, Esiobu, Fernandes, Fu, Fu, Fuller, Gao, Goswami, Goyal, Hartshorn, Hosseini, Hou, Inan, Kardas, Kerkez, Khabsa, Kloumann, Korenev, Koura, Lachaux, Lavril, Lee, Liskovich, Lu, Mao, Martinet, Mihaylov, Mishra, Molybog, Nie, Poulton, Reizenstein, Rungta, Saladi, Schelten, Silva, Smith, Subramanian, Tan, Tang, Taylor, Williams, Kuan, Xu, Yan, Zarov, Zhang, Fan, Kambadur, Narang, Rodriguez, Stojnic, Edunov, and Scialom]{llama2}
Hugo Touvron, Louis Martin, Kevin Stone, Peter Albert, Amjad Almahairi, Yasmine Babaei, Nikolay Bashlykov, Soumya Batra, Prajjwal Bhargava, Shruti Bhosale, Dan Bikel, Lukas Blecher, Cristian~Canton Ferrer, Moya Chen, Guillem Cucurull, David Esiobu, Jude Fernandes, Jeremy Fu, Wenyin Fu, Brian Fuller, Cynthia Gao, Vedanuj Goswami, Naman Goyal, Anthony Hartshorn, Saghar Hosseini, Rui Hou, Hakan Inan, Marcin Kardas, Viktor Kerkez, Madian Khabsa, Isabel Kloumann, Artem Korenev, Punit~Singh Koura, Marie-Anne Lachaux, Thibaut Lavril, Jenya Lee, Diana Liskovich, Yinghai Lu, Yuning Mao, Xavier Martinet, Todor Mihaylov, Pushkar Mishra, Igor Molybog, Yixin Nie, Andrew Poulton, Jeremy Reizenstein, Rashi Rungta, Kalyan Saladi, Alan Schelten, Ruan Silva, Eric~Michael Smith, Ranjan Subramanian, Xiaoqing~Ellen Tan, Binh Tang, Ross Taylor, Adina Williams, Jian~Xiang Kuan, Puxin Xu, Zheng Yan, Iliyan Zarov, Yuchen Zhang, Angela Fan, Melanie Kambadur, Sharan Narang, Aurelien Rodriguez, Robert Stojnic, Sergey Edunov, and Thomas
  Scialom.
\newblock Llama 2: Open foundation and fine-tuned chat models, 2023{\natexlab{b}}.

\bibitem[Wenlong et~al.(2023)Wenlong, Chen, Ruohan, Yunzhu, Jiajun, and Li]{voxposer}
Huang Wenlong, Wang Chen, Zhang Ruohan, Li Yunzhu, Wu Jiajun, and Fei-Fei Li.
\newblock Voxposer: Composable 3d value maps for robotic manipulation with language models, 2023.

\bibitem[Wu et~al.(2023)Wu, Yin, Qi, Wang, Tang, and Duan]{visual}
Chenfei Wu, Shengming Yin, Weizhen Qi, Xiaodong Wang, Zecheng Tang, and Nan Duan.
\newblock Visual chatgpt: Talking, drawing and editing with visual foundation models, 2023.

\bibitem[Wu et~al.(2019)Wu, Kirillov, Massa, Lo, and Girshick]{detectron2}
Yuxin Wu, Alexander Kirillov, Francisco Massa, Wan-Yen Lo, and Ross Girshick.
\newblock Detectron2.
\newblock \url{https://github.com/facebookresearch/detectron2}, 2019.

\bibitem[Zhang et~al.(2023)Zhang, Han, Qiao, Kim, Bae, Lee, and Hong]{faster}
Chaoning Zhang, Dongshen Han, Yu Qiao, Jung~Uk Kim, Sung-Ho Bae, Seungkyu Lee, and Choong~Seon Hong.
\newblock Faster segment anything: Towards lightweight sam for mobile applications, 2023.

\bibitem[Zhao et~al.(2022)Zhao, Zhang, Sun, Liu, Wang, and Yu]{yolonas}
Jianfeng Zhao, Lei Zhang, Jian Sun, Zilong Liu, Jingdong Wang, and Fisher Yu.
\newblock Yolo-nas: Neural architecture search for real-time object detection, 2022.

\bibitem[Zhao et~al.(2023)Zhao, Ding, An, Du, Yu, Li, Tang, and Wang]{fast}
Xu Zhao, Wenchao Ding, Yongqi An, Yinglong Du, Tao Yu, Min Li, Ming Tang, and Jinqiao Wang.
\newblock Fast segment anything, 2023.

\bibitem[Zhu et~al.(2023)Zhu, Chen, Shen, Li, and Elhoseiny]{minigpt}
Deyao Zhu, Jun Chen, Xiaoqian Shen, Xiang Li, and Mohamed Elhoseiny.
\newblock Minigpt-4: Enhancing vision-language understanding with advanced large language models, 2023.

\end{thebibliography}
}


\end{document}